Accepted Manuscript

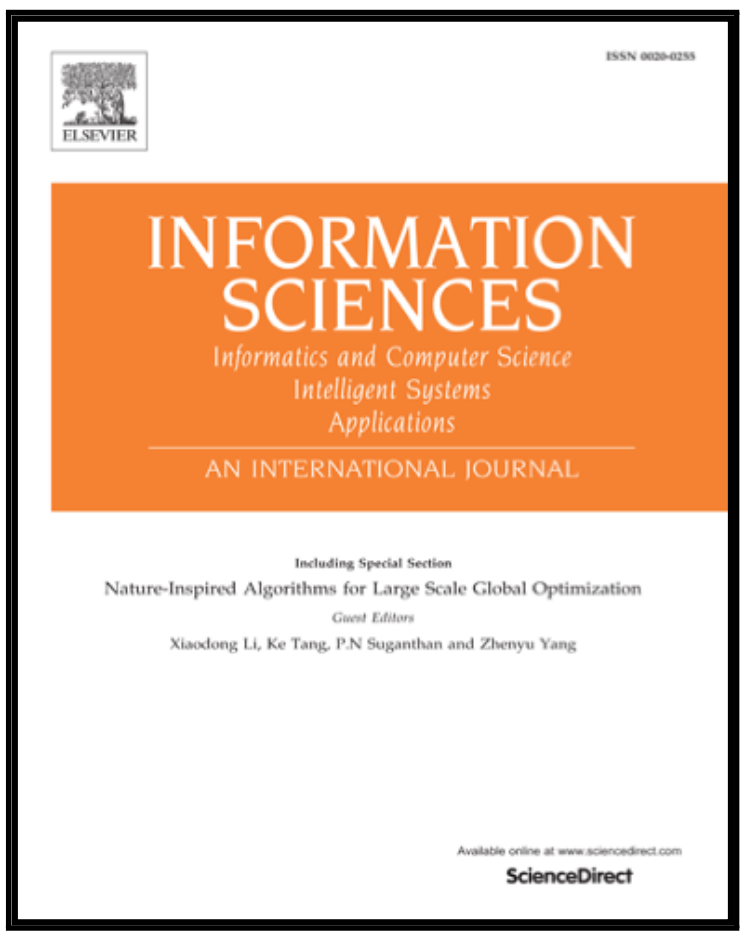

# Across neighbourhood search for numerical optimization

Guohua Wu

# Across neighbourhood search for numerical optimization

Guohua Wu*

*Science and Technology on Information Systems Engineering Laboratory, National University of Defense Technology, Changsha 410073, Hunan, P.R. China*

**Abstract:** Population-based search algorithms (PBSAs), including swarm intelligence algorithms (SIAs) and evolutionary algorithms (EAs), are competitive alternatives for solving complex optimization problems and they have been widely applied to real-world optimization problems in different fields. In this study, a novel population-based across neighbourhood search (ANS) is proposed for numerical optimization. ANS is motivated by two straightforward assumptions and three important issues raised in improving and designing efficient PBSAs. In ANS, a group of individuals collaboratively search the solution space for an optimal solution of the optimization problem considered. A collection of superior solutions found by individuals so far is maintained and updated dynamically. At each generation, an individual directly searches across the neighbourhoods of multiple superior solutions with the guidance of a Gaussian distribution. This search manner is referred to as across neighbourhood search. The characteristics of ANS are discussed and the concept comparisons with other PBSAs are given. The principle behind ANS is simple. Moreover, ANS is easy for implementation and application with three parameters being required to tune. Extensive experiments on 18 benchmark optimization functions of different types show that ANS has well balanced exploration and exploitation capabilities and performs competitively compared with many efficient PBSAs (Related Matlab codes used in the experiments are available from http://guohuawunudt.gotoip2.com/publications.html).

**Keywords:** Across neighbourhood search, evolutionary computation, swarm intelligence, numerical optimization

---

This work was partly supported by the National Natural Science Foundation of China under Grant No. 61563016, 41571397 and 51178193.

* Corresponding author Tel.: +86 15580845945

Email address: guohuawu@nudt.edu.cn

## 1. Introduction

Optimization, including continuous optimization and discrete optimization, or constrained optimization and unconstrained optimization, is frequently involved in many areas, ranging from engineering, management to commercial. Methods for solving optimization problems are referred to as optimization methods. Diverse mathematical programming methods [43], such as fast steepest, conjugate gradient method, quasi-newton methods, sequential quadratic programming, were first extensively investigated. However, increasing evidences have shown that these traditional mathematical optimization methods are generally inefficient or not efficient enough to deal with many real-world optimization problems characterized by being multimodal, non-continuous and non-differential [61].

In response to this challenge, many population-based search algorithms (PBSAs), including swarm intelligence algorithms (SIAs) and evolutionary algorithms (EAs), have been presented and demonstrated to be competitive alternative algorithms, such as the classical and popular genetic algorithm (GA) [15, 16], evolutionary programming (EP) [14, 67], particle swarm optimization (PSO) [12, 30], differential evolution (DE) [7, 49, 56] and ant colony optimization (ACO) [2, 10]. PBSAs is especially prominent in some optimization areas, such as multiobjective optimization [52, 53], multimodal optimization [6, 57] and complex constrained optimization [40, 54].

We have witnessed that PBSAs have progressed continuously in recent years while gaining great success in real-world applications. However, according to the No Free Lunch (NFL) theorems [58], all search algorithms, including PBSAs, will averagely possess the same performance when they are applied to all possible optimization problems, that is to say, theoretically, there will not exist a general optimization algorithm being superior to all other algorithms. As a result, in addition to the extensive studies on classical PBSAs, new PBSAs with specialized principles and search strategies are emerging more recently to provide more choices for users, such as artificial bee colony (ABC) [1, 28], biogeography-based optimization (BBO) [37, 48], chemical reaction optimization (CRO) [31, 32] and a group search optimizer (GSO) [20].

Roughly speaking, three directions might be deserved to give attention to in order to prompt the progress of evolutionary and swarm intelligence computation. The first direction is associated with the sophisticated modification of existing PBSAs to get performance-enhanced algorithm versions. In fact, we can find that currently, major works on PBSAs are in line with this direction, including the hybridization of different PBSAs [17, 26], adaptive parameter control [3, 68] and intelligent combination of different search strategies [22, 34, 60]. The reasonable balance between intensification and diversification, or exploitation and exploration is crucial to improve the efficiency of PBSAs [63].

The second direction attracting researchers' awareness involves the effective integration of problem domain knowledge into PBSAs. Although PBSAs generally do not rely on specific problem domain knowledge, which enables them to be suited to diverse optimization problems, evidences show that the appropriate combination with domain knowledge could often significantly strengthen the performance of PBSAs when dealing with specific problems. For example, the proper use of domain knowledge in ACO application can facilitate to more effective solution representation, neighborhood construction and search strategy design [11, 59, 65]. In addition, domain knowledge related gradient information and variable relationships were employed in PBSAs for continuous optimization [44, 61, 64]. Recently, Wu et.al proposed an equality constraint and variable reduction strategy (ECVRS) by employing variable relationships to reduce equality constraints as well as variables of constrained optimization problems [62].

The third direction is the design of new PBSAs. As mentioned before, although various PBSAs have been proposed, the NFL theorems tell that any algorithm cannot be efficient for all optimization problems. To deal with vast number of optimization problems encountered in the real world, new PBSAs with effective and unique optimization strategies are still needed. Generally, to guarantee the contribution of a new PBSA, three standards should be satisfied. Firstly, the principles and concepts behind the new PBSA should be different from other PBSAs. Secondly, the mechanisms included in the new PBSA should be simple, understandable and easy for application. Thirdly and more importantly, the new PBSA should be better than or at least as competitive as recent popular and efficient PBSA variants.

In this study, the author proposes a novel population-based optimizer named across neighbourhood search (ANS). Like other swarm intelligence algorithms (e.g., PSO and ACO), in ANS, a group of individuals search in the solution space with the aim to find the optimal solution of an optimization problem. A memory collection is utilized in ANS to record a certain number of superior solutions found so far by the whole population. At every generation, each individual updates its position by searching across the neighbourhoods of multiple superior solutions biased by Gaussian distribution. ANS is very easy and convenient for implementation and application with three parameters requiring adjustments to cater for different optimization problems. Moreover, extensive experiments on various benchmark functions, including unimodal, multimodal and rotated functions, demonstrate that the overall performance of ANS is very competitive compared with several peer PBSAs.

The rest of the paper is structured as follows. Section 2 introduces the new proposed across neighbourhood search (ANS), including its motivations, search strategies, convergence process, and algorithmic framework. Section 3 comprehensively discusses the differences between ANS and other major PBSAs. Section 4 reports the experimental results and comparative studies of ANS. Section 5 analyzes the impacts of parameters of ANS. Section 6 concludes this study and gives directions of future research.

## 2. Across neighbourhood search

### 2.1. Motivations of ANS

Learning from better solutions or individuals is a common technique in PBSAs, though the processes to realize such learning mechanism could be exhibited in different ways for different PBSAs. For example, in PSO, a particle flies in the solution space with a velocity that is guided by the local best solution and global best solution. The learning mechanism is realized during the process that particles fly to local or global best solutions. In GA, EP and DE, at each generation the selection operation is used to retain better solutions while the related crossover and mutation operations based on the retained better solutions actually realize the learning mechanism between better solutions. Although without explicit declaration, these better-solution based learning mechanisms in different PBSAs are all on the basis of following two assumptions.

1) It has a higher probability to find another better solution around a superior solution in the solution space.

2) High-quality solutions share similarities with the theoretical optimal solution, namely they have some good solution components (good values for some variables or dimensions, for example).

If the assumptions above hold, to design an effective PBSA, we need to give satisfactory answers to following three questions.

1) How to retain and update a collection of superior solutions for all individuals in a population? To deal with multimodal optimization problems, multiple superior solutions are also expected to be beneficial to the location of multiple local optima.

2) How to search around (or learn from) a superior solution for each individual? Learning from superior solutions is very important to guarantee the convergence of an individual, since better solutions are expected to locate around a known superior solution, as indicated in assumption 1.

3) For each individual, how to learn from multiple superior solutions simultaneously to get potential good solution components from them? This is because assumption 2 tells that all superior solutions potentially imply the knowledge about reaching the theoretical optimal solution. This question is also very important for two reasons. First of all, it can help individuals to comprehensively learn the correlation knowledge implied in different superior solutions. Secondly, it can effectively prevent individuals from the premature convergence to one local optimum.

### 2.2. The search strategy of ANS

In this section, ANS is described along with giving answers to the questions proposed in Section 2.1. For the sake of clear description, some notations are firstly defined. Suppose that a group of $m$ individuals search in a solution space. Let $pos_i$ denote the current position of individual $i$; Collection $R$ includes a certain number of superior solutions; Each superior solution is denoted by $r_i$; The cardinality of $R$ is denoted by $c$.

Corresponding to the three issues raised in the former section, three main parts are included in ANS.

1) Maintenance of the superior solution collection $R$

In the current ANS version, the best solution found by each individual $i$ so far is taken as a superior solution $r_i$. As a result, the cardinality of the superior solution collection is equal to the population size of ANS, i.e. $c$ is set to $m$ (i.e. $c = m$). Although such setting is only a special case of ANS, as $c$ can be larger or smaller than $m$, experiments show that it helps ANS to have satisfactory performance in solving different types of optimization problems. More sophisticated configuration approach for $c$ (e.g. self-adaption) will be investigated in the future study.

2) Search of the neighbourhood of a superior solution

As the current best so far solution $r_i$ (i.e. position) of each individual $i$ is taken as one superior solution, individual $i$ is asked to search the neighbourhood of solution $r_i$. At every generation, each individual $i$ searches an approximate hyper-box area determined by its current position $pos_i$ and the superior position $r_i$. This hyper-box area is just the neighbourhood of $r_i$. The center of the hyper-box is $r_i$. The search range for $d$ th dimension of individual $i$ is illustrated in Fig. 1, from which we can observe that $r_i^d$ is the center of the search range and $|r_i^d - pos_i^d|$ is the related approximate semi-length. To realize the random search in this range, the Gaussian distribution function is employed, which means that the value closer to $r_i^d$ has a relatively higher probability to be assigned to $pos_i^d$. As a result, each individual is able to explore the whole neighbourhood while tends to search the region closer to the superior solution (the center), which is consistent with the assumption 1. The mathematical formulation for the position update strategy of individual $i$ when searching the neighbourhood of $r_i$ is given in (1), where $G(0,\sigma^2)$ is a Gaussian random value with mean value of zero and standard deviation of $\sigma$.

$$pos_i^d = r_i^d + G(0,\sigma^2) * |r_i^d - pos_i^d| \quad (1)$$

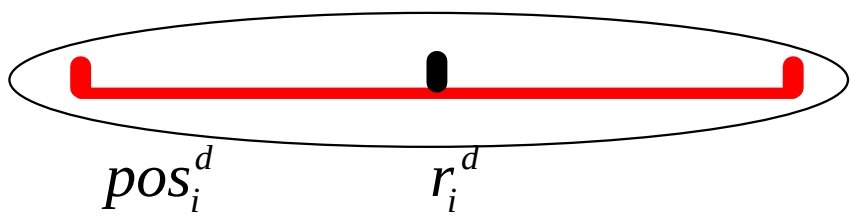

Fig. 1 Illustration of the search range on one dimension

3) Simultaneous search of the neighbourhoods of other superior solutions

Let $D$ be the dimensionality of a numerical optimization problem considered. To enable individual $i$ to search across the neighbourhoods of other superior solutions simultaneously, $n$ dimensions of $pos_i$ are randomly selected ($n$ is referred to as across-search degree). $N$ denotes the collection that includes the $n$ ($0 \le n \le D$) selected dimensions. For each randomly selected dimension $d$ in $N$, a randomly selected superior solution $r_{g(d)}$ ($g(d) \ne i$) is used to replace $r_i$ to realize the search on dimension $d$. The values of dimensions in set $N$ are updated as

$$pos_i^d = r_{g(d)}^d + G(0,\sigma^2) * | r_{g(d)}^d - pos_i^d |, \text{ for } d \in N. \tag{2}$$

Combing (1) and (2), the position update rule of individual $i$ is described as

$$pos_i^d = \begin{cases} r_i^d + G(0,\sigma^2) * | r_i^d - pos_i^d | & \text{if } d \notin N \\ r_{g(d)}^d + G(0,\sigma^2) * | r_{g(d)}^d - pos_i^d |, \ g(d) \ne i & \text{otherwise } d \in N \end{cases}. \tag{3}$$

From (3), we can find that each individual in ANS can search across the neighbourhoods of multiple superior solutions. This search strategy is exactly referred to as across neighbourhood search in this study.

The scalar parameter $\sigma$ impacts the exploration and exploitation capabilities of ANS. It can be easily perceived that a larger $\sigma$ enables individuals to explore larger areas, while smaller $\sigma$ enables individuals to focus their search efforts on smaller regions closer to superior solutions. For a Gaussian random variable $x \sim G(0,\sigma^2)$, we have the following probability distribution

$$P(-\sigma < x < \sigma) = 0.6826, \tag{4}$$

$$P(-2\sigma < x < 2\sigma) = 0.9544. \tag{5}$$

According to (4) and (5), there is a probability of 0.6826 for a dimension $d$ of individual $i$ to be assigned a value among $[r_i^d - \sigma * | r_i^d - pos_i^d |, r_i^d + \sigma * | r_i^d - pos_i^d |]$ and 0.9544 to be assigned a value among $[r_i^d - 2\sigma * | r_i^d - pos_i^d |, r_i^d + 2\sigma * | r_i^d - pos_i^d |]$. To balance the intensification and diversification of ANS, each individual $i$ should not spend too much effort in searching the space out of the range of $[r_i^d - | r_i^d - pos_i^d |, r_i^d + | r_i^d - pos_i^d |]$. In addition, individuals need to explore this range sufficiently at the earlier stage of the algorithm. Therefore, a proper value of parameter $\sigma$ could be 0.5. In this case, there is a probability of 0.9544 that individual $i$ searches in the range of $[r_i^d - | r_i^d - pos_i^d |, r_i^d + | r_i^d - pos_i^d |]$.

The general framework of the proposed ANS is described in Algorithm 1.

**Algorithm 1: The framework of ANS**

Initialize the population size $m$, cardinality $c$ of superior solution set, across-search degree $n$, standard deviation $\sigma$ of the Gaussian distribution, and problem dimensionality $D$;
Randomly initialize the position $pos_i$ of each individual $i$;
Set the superior solution $r_i$ to $pos_i$;
Record the best solution $g$ found by the whole population so far;
Initialize the allowed maximum generations $MaxG$ and set the current generation $k=1$;
**While** $k < MaxG$
    $k = k+1$**;**
    **For** $i = 1 \rightarrow m$
        Let set $N$ record the randomly selected $n$ $(1 \le n \le D)$ dimensions for individual $i$;
        **For** $d = 1 \rightarrow D$ **If** $d \notin N$
            $pos_i^d = r_i^d + G(0,\sigma^2) * | r_i^d - pos_i^d |$;
          **Else if** $d \in N$
            Randomly select a superior solution $r_{g(d)}$ from $R$ $(g(d) \neq i)$;
            $pos_i^d = r_{g(d)}^d + G(0,\sigma^2) * | r_{g(d)}^d - pos_i^d |$;
          **End if**
        **End for**
        **If** $pos_i$ is better than $r_i$
          Update $r_i$ with $pos_i$;
        **End if**
        **If** $pos_i$ is better than $g$
          Update $g$ with $pos_i$;
        **End if**
    **End for**
**End while**

It is worth noting that Gaussian distribution is a popular technique that facilitates to the random search behaviour of EAs and SIAs. For example, conventional EP utilizes a normal Gaussian distribution to realize the mutation operator [67], and Gaussian distribution mutation strategy has been used to maintain the diversity of PSO [21, 47]. Recently, compact EAs attract an increasing attention. Several compact EAs have been proposed, such as compact DE[41], compact PSO [42] and compact (bacterial foraging optimization) BFO [24]. The main feature of compact EAs is that they are population-less. Instead of processing an actual population of solution, a virtual population is used. Gaussian distribution is generally employed in generating the virtual population. Similar to Gaussian distribution, other distributions like Cauchy distribution [67] and Uniform distribution [23] are also exploited in literature. The Gaussian distribution guided search behavior of ANS is different from previous studies in two aspects. First, it is scaled by the length between the current solution and one superior solution in each dimension, which may enable ANS to have good exploitation and exploration capability. Second, it searches across neighborhoods of multiple superior solutions, thus prompting ANS to quickly learn high-quality components from more promising solutions and preventing ANS from converging to local optima prematurely.

### 2.3 Graphical illustration of the convergence of ANS

In this section, the position distributions of individuals and superior solutions are plotted when using ANS to solve the Rastrigin function with two dimensions. Rastrigin is known as a representative multimodal function with many local optima. The convergence states (i.e. position distributions of individuals and superior solutions) of ANS at generation 10, 20, 40, 60 and 80 are displayed in Fig. 2.

Several observations can be obtained from Fig. 2. First, ANS shows satisfactory convergence in dealing with the Rastrigin function. Actually, it finds the global best solution within 80 generations. Second, ANS is able to locate multiple local optima when searching the landscape of Rastrigin (refer to the distributions of superior solutions in Fig. 2b1 and Fig. 2b2 as examples). The recorded superior solutions gradually converge to the global optimum as ANS proceeds. Third, superior solutions generally locate in landscape basins while individuals are capable of exploring unvisited regions.

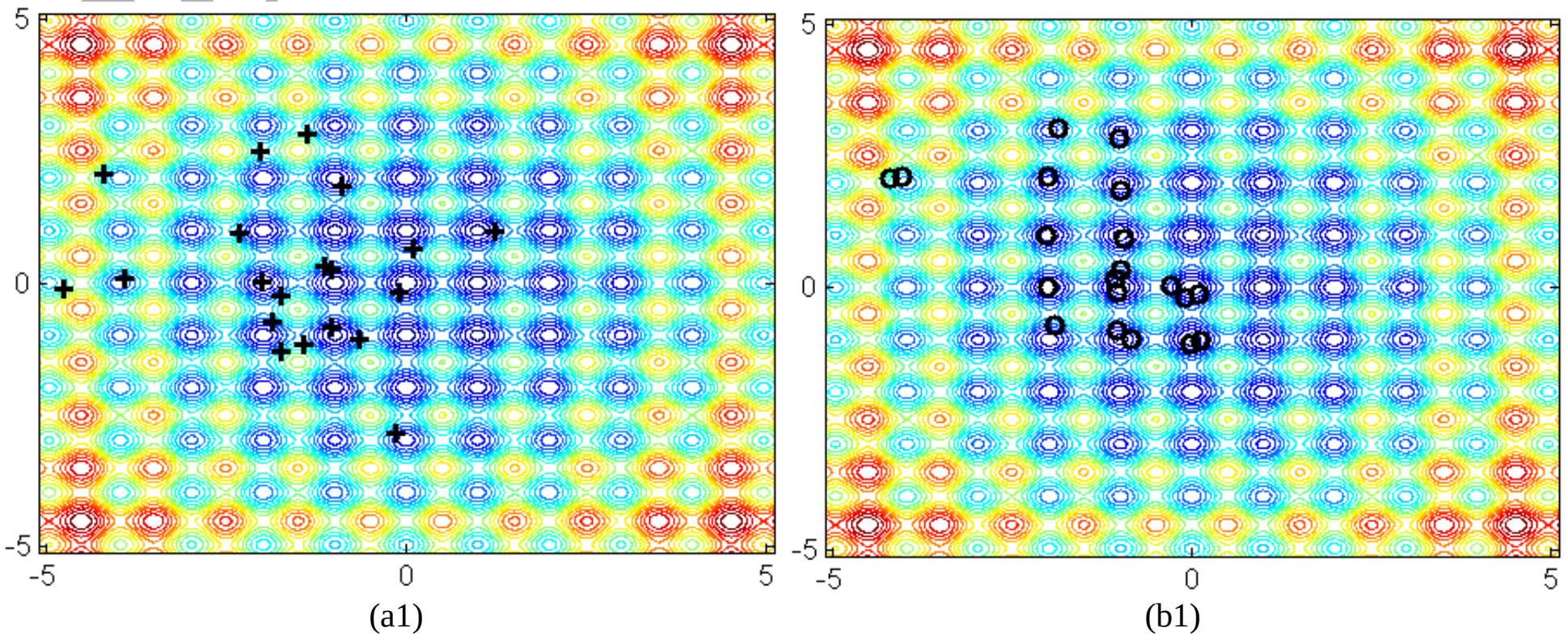

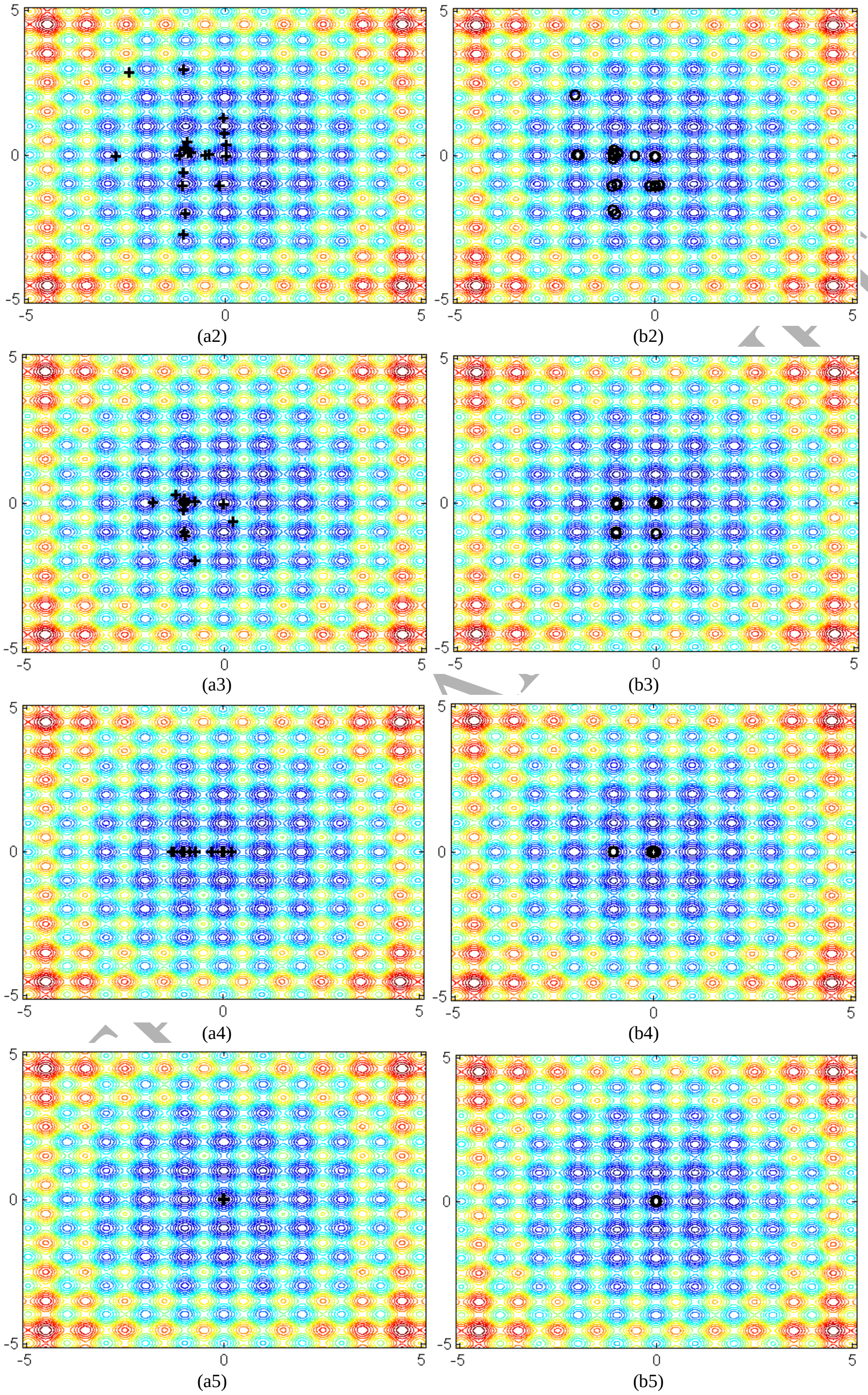

Fig. 2 Illustration of the convergence process of ANS in solving Rastrigin function with two dimensions. Related parameters used in ANS are set to $m = c = 20$, $n = 1$ and $\sigma = 0.5$. Subfigure (a1), (a2), (a3), (a4) and (a5) show the position distribution of individuals at generation 10, 20, 40, 60 and 80, respectively. Subfigure (b1), (b2), (b3), (b4) and (b5) display the superior solution distribution at generation 10, 20, 40, 60 and 80, respectively.

### 2.4 Characteristics of ANS

In contrast to some other peer PBSAs, the proposed ANS algorithm exhibits some special characteristics.

1) In ANS the search strategies and interaction mechanisms among individuals are very simple and easy to understand. The idea implied in ANS is also very straightforward: individuals perform the Gaussian distribution guided across neighbourhood search, which is different from current existing PBSAs.

2) Only three parameters are required to tune to cater for different optimization problems, i.e. cardinality (denoted by $c$) of the superior solution set, the across-search degree (denoted by $n$), and the standard deviation (denoted by $\sigma$) of the Gaussian distribution. In general, the cardinality ($c$) can be equal to the overall number ($m$) of individuals in the population. Guidance on the parameter setting of ANS will be introduced in Section 5.

3) It is very easy and convenient to implement and apply ANS. Our experience shows that to realize ANS, only a few lines of Matlab codes can work.

It can be perceived that ANS is very simple and the adopted search strategy and parameters in ANS are well motivated. Actually, in [25], Iacca et al have shown that a very simple algorithm, if properly designed, can outperform much more complex and computationally expensive approaches. They further argued that attention should be paid to the algorithmic design phase which should be accompanied by an algorithmic philosophy, i.e. the presence of each element should be intuitively understood and justified. In other words, the law of parsimony by William Ockham (known as Ockham razor) would be an appropriate philosophical guidance and source of inspiration for designing efficient optimization algorithms.

Conceptual and experimental analyses are given in Section 3 and Section 4, respectively.

## 3. Conceptual comparisons with other PBSAs

Before proceeding to the conceptual comparisons between ANS and other classical PBSAs, the search behaviours of ANS, canonical PSO and canonical DE in a two dimension space are illustrated in Fig. 3, Fig. 4 and Fig. 5, respectively, providing readers an intuitive impression on the search manner of ANS and the differences from other popular PBSAs. Readers are supposed to be familiar with PSO and DE. The characteristic of one-step search process of each PBSA considered can be clearly captured from Fig. 3, Fig. 4 and Fig. 5. Intuitively, ANS do have a different search principle. In addition, it can be observed that the number of solutions (also vectors or positions) required in the update process are 3, 4 and 5 in ANS, PSO and DE, respectively.

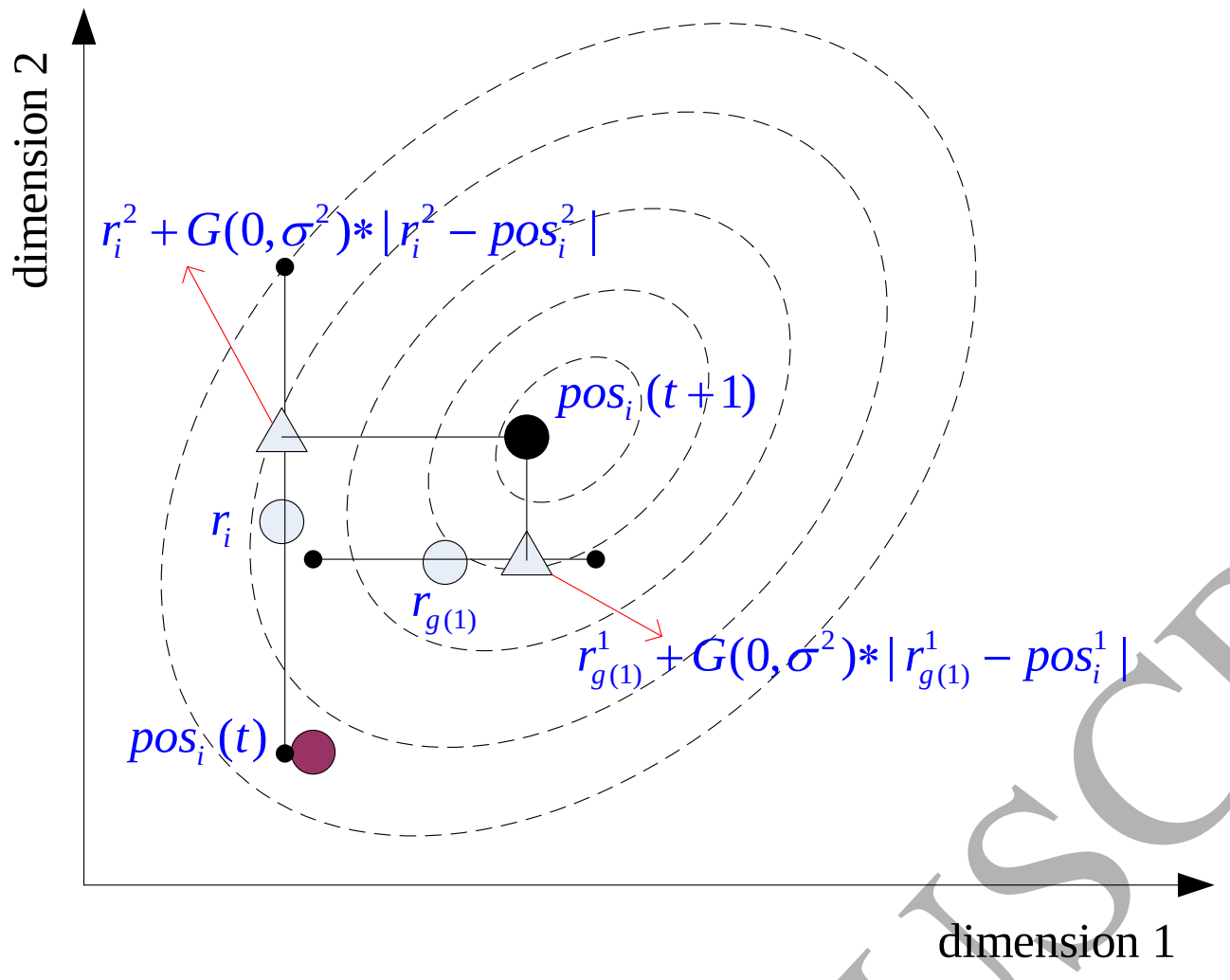

Fig. 3 Illustration of the search behavior of ANS in 2-D solution space. The $i$th individual learns from different superior solutions with respect to different dimensions. In dimension one, the position is updates to $r_{g(1)}^1 + G(0,\sigma^2) * | r_{g(1)}^1 - pos_i^1 |$ while in dimension two it is updated to $r_i^2 + G(0,\sigma^2) * | r_i^2 - pos_i^2 |$, such that the new position $pos_i(t+1)$ is obtained.

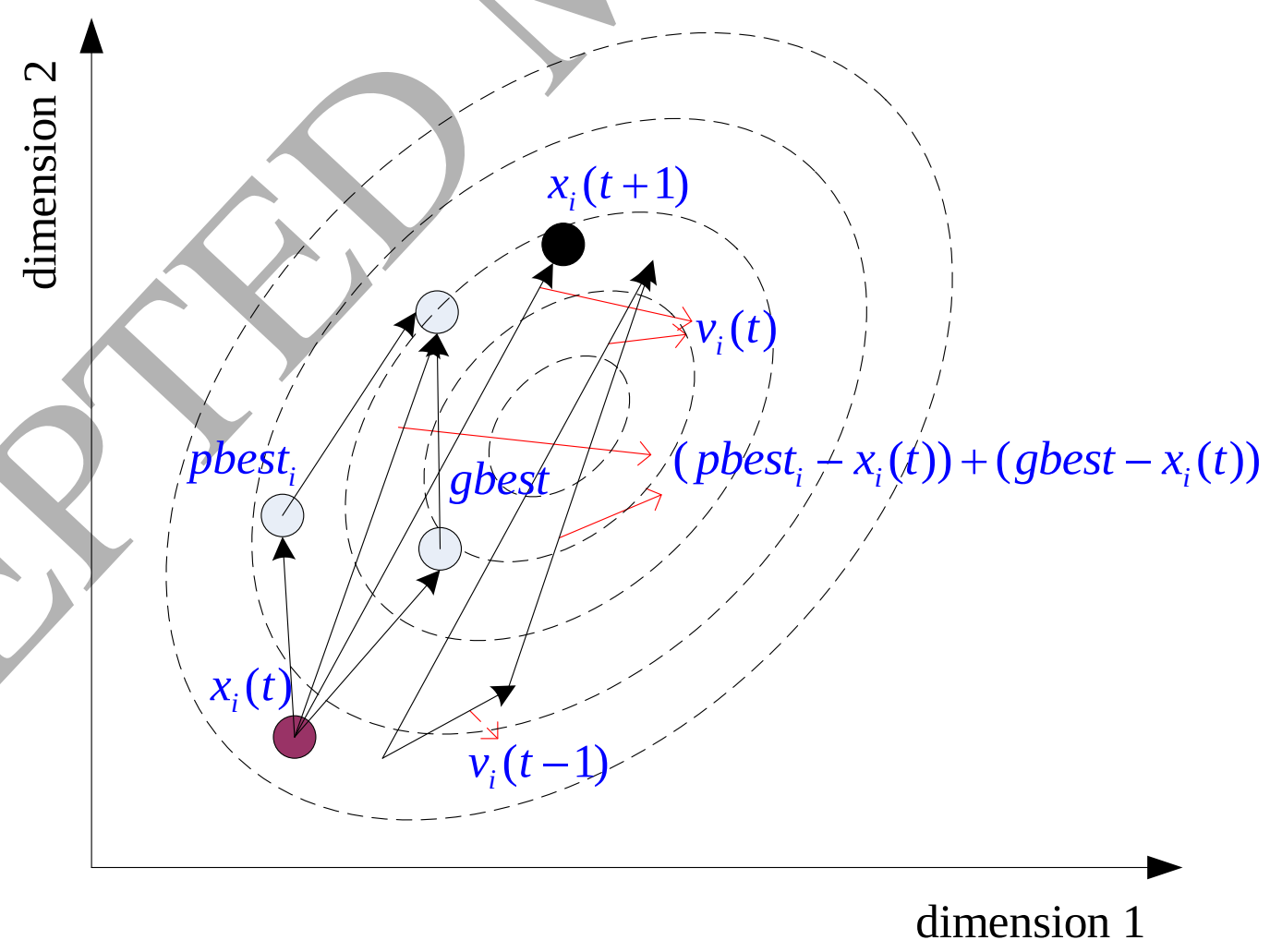

Fig. 4 Illustration of the search behavior of classical PSO in 2-D solution space. Notation $t$ in parentheses denotes the current generation. The position of the $i$th particle is updated as: $v_i(t) = w \times v_i(t-1) + c_1 \times r_1 \times (pBest_i - x_i(t)) + c_2 \times r_2 \times (gBest - x_i(t))$ and $x_i(t+1) = x_i(t) + v_i(t)$, where $x_i(t)$ and $v_i(t)$ are the position and velocity of the $i$th particle. $pBest_i$ and $gBest$ denote the local best and global best positions, respectively. $w$ is a parameter called inertial weight, and $c_1$ and $c_2$ are acceleration parameters. For the sake of simplicity, parameters $w$, $c_1$ and $c_2$ are removed in the plot. $r_1$ *and* $r_2$ *are random values among [0,1].*

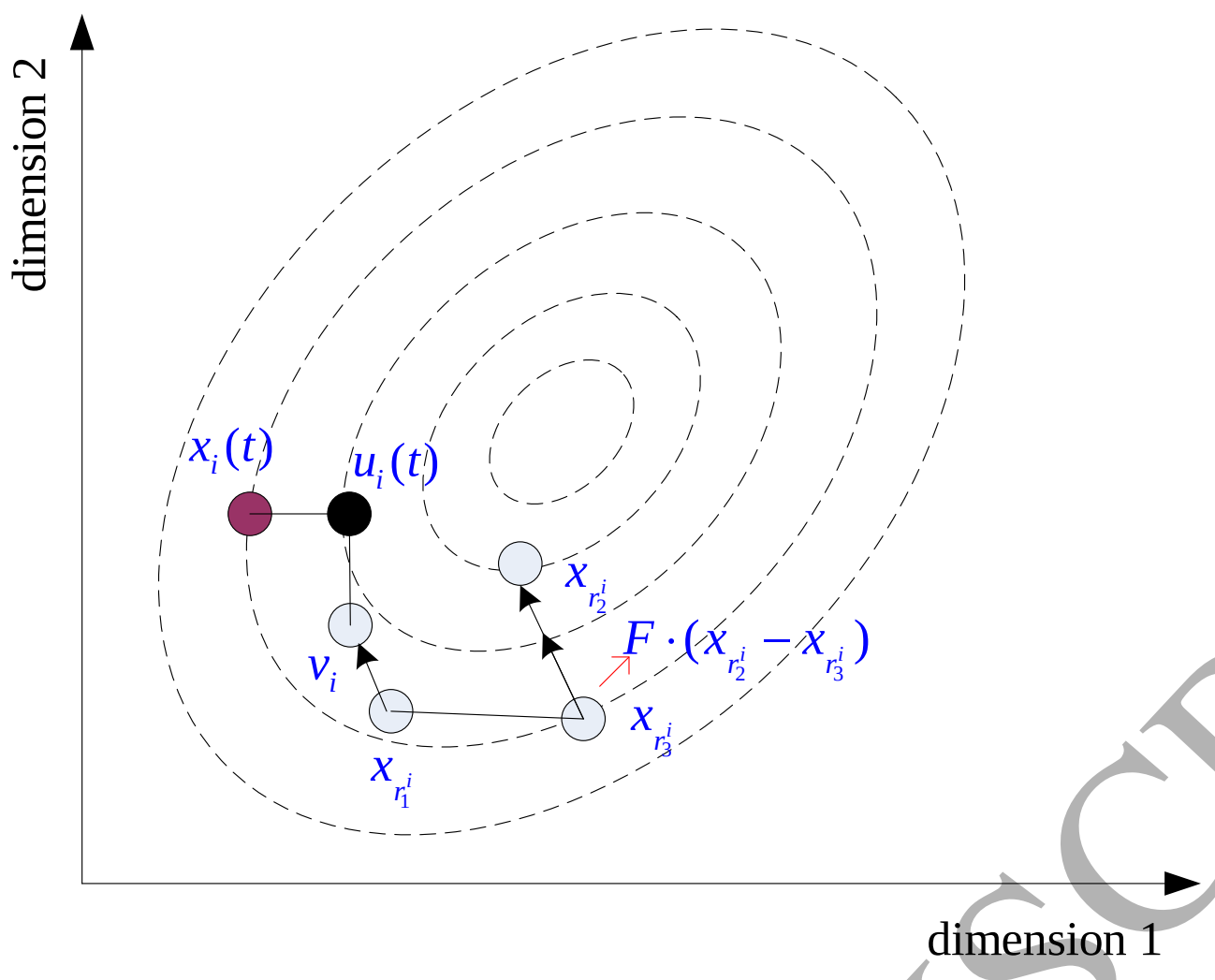

Fig. 5 Illustration of the search behavior of classical PSO in 2-D solution space. The popular "DE/rand/1/bin" is illustrated here. The mutant vector of $x_i$ is calculated by the mutation operator $v_i = x_{r_1^i} + F.(x_{r_2^i} - x_{r_3^i})$. Afterwards, the related trial vector is obtained through a binomial crossover operator, $u_i^j = \begin{cases} v_i^j & \text{if } (rand_j[0,1] \le CR) \text{ or } (j = j_{rand}), j = 1, 2, ..., D \\ x_i^j & \text{otherwise} \end{cases}$. The better one between $x_i$ and $u_i$ will survive to next generation.

As GA, DE and PSO are among the most popular PBSAs for numerical optimization currently, the conceptual comparisons between ANS and GA, DE and PSO are given in this section. Although some existing PBSAs are analogous to certain natural phenomena (e.g. GA is inspired by the natural evolution process and PSO mimics the behaviours of a flocking of birds) while some are not, they all share some common features. Generally speaking, two steps are shared by most PBSAs, i.e. to produce new solutions and to evaluate the new solutions at every generation (or iteration). In addition, in the process of solution generation and evaluation, corresponding techniques are employed in each PBSA to balance the intensification and diversification (or exploration and exploitation). As a result, we may come to a conclusion that PBSAs differ from each other in their strategies adopted to realize the solution generation, solution evaluation, exploration and exploitation. The differences between ANS and other PBSAs are discussed from these four aspects.

As for solution generation, in PSO, the position (positions represent solutions in PSO) of a particle is updated by moving forward in terms of its current velocity. The current velocity of the particle is determined by its former velocity and the attraction of the global best position and its own or its neighbours' local best positions. In GA and DE, new solutions are generated by crossover and mutation operations. One difference between GA and DE is that GA uses chromosomes (a chromosome denotes a solutions in GA) to crossover directly to produce new chromosomes and then a mutation operation is exerted on the newly produced chromosomes to increase the diversity of population, while DE uses mutation operation to generate donor vectors (in DE, a vector records a solution) then the crossover operation is performed between target vectors and donor vectors. In contrast, in ANS, an individual directly searches across the neighbourhoods of multiple superior solutions to produce a new solution. The size and position of the neighbourhood are determined by the current position of the individual and the superior solutions. Therefore, from the perspective of solution generation, the principle behind ANS is simpler and different from PSO, GA and DE.

With regard to solution evaluation, in GA and DE, selection operations are adopted to choose proper high-quality individuals to survive for next generation, which could be deterministic or stochastic. In PSO, new positions will be evaluated to try to update local best positions and the global best position. In contrast, in ANS, the new positions of individuals will be evaluated to determine whether they are suited to update the superior solution collection.

Exploration and exploitation is a crucial pair of concepts in improving and designing PBSAs, whereas they usually contradict with each other in the evolutionary process of PBSAs. Črepinšek et al. comprehensively surveyed the exploration and exploitation related topics in EAs [4]. They defined that **exploration** is the process of visiting entirely new regions of a search space and whilst **exploitation** is the process of visiting those regions of a search space within the neighbourhood of previously visited points [4]. Moreover, useful exploration and exploitation measures have been investigated and utilized to guide the parameter control of EAs [36].

Generally, to avoid being trapped in local optima and suffering from premature convergence, a strong exploration capability may be needed. On the contrary, to enhance the local search ability and converge to a better solution more quickly, it is necessary to strengthen the exploitation capability. In PSO, the sustainable attraction of local best and global best positions enables particles to have promising exploitation capabilities. The random values and accelerate coefficients utilized to update the velocities of particles are beneficial to the exploration capabilities of particles. In DE and GA, the crossover operator potentially can spread high-quality solution components into the population and the selection operator supports the survival of high-quality solutions. Therefore, the crossover and selection operators are important to DE and GA to possess satisfactory exploitation capability. On the contrary, the mutation operator in DE and GA can maintain the diversity of the population effectively which gives these two algorithms good exploration ability. Moreover, the crossover degree between solutions in DE and GA is generally also important to adjust the exploration and exploitation capabilities. Compared with GA, DE uses run-time solution differences to complete the mutation operation, which might be more useful and contributes to the high efficiency of DE in dealing with numerical optimization problems. Although in the canonical PSO, DE and GA versions, both exploration and exploitation related mechanisms have been implicitly included, experimental evidences have shown that their original versions often face the common problem of premature convergence. Therefore, many strategies related to parameter adaptation, topology adjustment, hybridization and coevolution have been incorporated into the conventional PBSAs to obtain performance-enhanced PBSA variants.

In comparison with PSO, DE and GA, ANS applies two mechanisms to enhance the exploration capabilities of individuals and prevent the premature convergence. The first mechanism is that the across neighbourhood search behaviour of an individual is determined by Gaussian random values and distances between the individual’s position and the superior solutions, which enable this individual to explore promising areas around superior solutions. The second exploration mechanism is that one individual does not focus all its search efforts on one superior solution. A predefined across-search degree (the number of selected dimensions) enable an individual to search across neighbourhoods of multiple superior solutions simultaneously, thus the individual can effectively make use of comprehensive knowledge from multiple superior solutions. This mechanism naturally is beneficial to preventing individuals from prematurely converging to one local optimum. The exploitation capability of ANS is reflected by following three aspects. First, the Gaussian random values guide individuals to search the areas closer to the superior solutions with a higher probability, which prompts individuals to move toward superior solutions. Second, the high-quality ones of the newly produced solutions are used to update the superior solution set. Third, along with the evolutionary process, the distance between the position of an individual and a superior solution will decrease in nature, which enables the individual to gradually have stronger exploitation capability at the later evolutionary stage.

Above analyses show that the new proposed ANS has its unique strategy in solution generation, solution evaluation and balance mechanisms for exploration and exploitation. Experimental verification of the performance of ANS will be given in Section 4.

## 4. Numerical experiments

### 4.1. Experimental comparison with popular and efficient PSO and DE variants

Currently, DE, PSO and ABC are three of the most popular PBSAs for numerical optimization. Therefore, in order to effectively justify the performance of ANS, we compare it with some popular and efficient PSO, DE and ABC variants as listed below.

**JADE:** adaptive differential evolution with an external archive [69];

**jDE:** differential evolution with self-adapting control parameters [3];

**CLPSO:** comprehensive learning particle swarm optimizer [35];

**FIPS:** fully informed particle swarm [38];

**FDR:** fitness-distance-ratio based particle swarm optimization [45];

**CPSO:** cooperative particle swarm optimization [50];

**ABC:** artificial bee colony [27, 29].

The parameters of the selected comparative algorithms above are the same as those suggested in the original papers. Related parameter settings are summarized in Table 1.

Table 1. Parameter configurations of comparative algorithms

| Algorithm | Parameter configuration |
|---|---|
| JADE | $p=0.05$, $c=0.1$, $NP$ is set to 100 and 400 in the case of $D=30$, and $=100$ |
| jDE | $\tau_1=\tau_2=0.1$, $F_l=0.1$, $F_u=0.9$, $NP=100$ |
| CLPSO | $w_o=0.9$, $w_1=0.4$, $c=1.49445$, $ps=40$, $Pc_i=0.05+0.45*(\exp(\frac{10(i-1)}{ps-1})-1)\Big/(\exp(10)-1)$ |
| FIPS | $w=0.7298$, $c_1=c_2=2$, $ps=30$ |
| FDR | $\varphi_1=1$, $\varphi_2=1$, $\varphi_3=2$, $w^{(i+1)}=(w^{(i)}-0.4)\times(gsize-i)/(gsize+0.4)$, $ps=30$ |
| CPSO | $c_1=c_2=1.49$, group number is equal to dimensionality, $w^{(i+1)}=0.9-i*0.5/gsize$, $ps=30$ |
| ABC | $limit=100$, colony size $NP=50$, the number of food sources is $NP/2$ |

Table 2. Benchmark optimization functions, $\mathbf{M}$ denotes the rotation matrix.

| Type | Function formula | Name | Search range |
|---|---|---|---|
| Uni-modal | $f_1(x)=\sum_{i=1}^{n} x_i^2$ | Sphere | [-500,500] |
| | $f_2(x)=\sum_{i=1}^{n}(100(x_i^2-x_{i+1})^2+(x_i-1)^2)$ | Rosenbrock | [-2.048, 2.048] |
| | $f_3(x)=\max\{\lvert x_i\rvert, 1\le i\le n\}$ | Schwefel 2.21 | [-10,10] |
| | $f_4(x)=\sum_{i=1}^{n}\lvert x_i\rvert+\prod_{i=1}^{n}\lvert x_i\rvert$ | Schwefel 2.22 | [-10,10] |
| | $f_5(x)=\sum_{i=1}^{n}\lfloor x_i+0.5\rfloor^2$ | Step function | [-100,100] |
| | $f_6(x)=\sum_{i=1}^{n} i x_i^4+rand[0,1)$ | Noise Quadric | [-2.048, 2.048] |
| Multi-modal | $f_7(x)=\sum_{i=1}^{n}(x_i^2-10\cos(2\pi x_i)+10)$ | Rastrigin | [-5.12, 5.12] |
| | $f_8(x)=\sum_{i=1}^{n}(y_i^2-10\cos(2\pi y_i)+10)$, $y_i=\begin{cases} x_i, if\ \lvert x_i\rvert<0.5 \\ 0.5*round(2*x_i), otherwise \end{cases}$ | Noncontinues Rastrigin | [-600, 600] |
| | $f_9(x)=-20\cdot\exp(-0.2\sqrt{\frac{1}{n}\sum_{i=1}^{n}x_i^2})+20-\exp(\frac{1}{n}\sum_{i=1}^{n}\cos(2\pi x_i))+e$ | Ackley | [-32, 32] |
| | $f_{10}(x)=\frac{1}{4000}\sum_{i=1}^{n}x_i^2+1-\prod_{i=1}^{n}\cos(\frac{x_i}{\sqrt{i}})$ | Griewank | [-600, 600] |
| | $f_{11}(x)=\frac{\pi}{D}\{10\sin^2(\pi y_1)+\sum_{i=1}^{D-1}(y_i-1)^2\times[1+10\sin^2(\pi y_{i+1})]+(y_D-1)^2\}+\sum_{i=1}^{D}u(x_i,10,100,4)$ $y_i=1+\frac{1}{4}(x_i+1)$, $u(x_i,10,100,4)=\begin{cases} k(x_i-a)^m, & x_i>a \\ 0, & -a\le x_i\le a \\ k(-x_i-a)^m, & x_i<-a \end{cases}$ | Penalized 1 | [-50,50] |
| | $f_{12}(x)=0.1\{\sin^2(3\pi x_1)+\sum_{i=1}^{D-1}(x_i-1)^2\times[1+\sin^2(3\pi x_{i+1})]+(x_D-1)[1+\sin^2(3\pi x_D)]\}+\sum_{i=1}^{D}u(x_i,5,100,4)$ | Penalized 2 | [-50,50] |
| Rotated | $f_{13}(x)=\sum_{i=1}^{n}z_i^2, \mathbf{z}=\mathbf{M}\cdot\mathbf{x}$ | Rotated Sphere | [-500,500] |
| | $f_{14}(x)=\sum_{i=1}^{n}(100(z_i^2-x_{i+1})^2+(z_i-1)^2), \mathbf{z}=\mathbf{M}\cdot\mathbf{x}$ | Rotated Rosenbrock | [-2.048, 2.048] |
| | $f_{15}(x)=\max\{\lvert z_i\rvert, 1\le i\le n\}, \mathbf{z}=\mathbf{M}\cdot\mathbf{x}$ | Rotated Schwefel 2.21 | [-10,10] |
| | $f_{16}(x)=\sum_{i=1}^{n}(z_i^2-10\cos(2\pi z_i)+10), \mathbf{z}=\mathbf{M}\cdot\mathbf{x}$ | Rotated Rastrigin | [-5.12, 5.12] |
| | $f_{17}(x)=-20\cdot\exp(-0.2\sqrt{\frac{1}{n}\sum_{i=1}^{n}z_i^2})+20-\exp(\frac{1}{n}\sum_{i=1}^{n}\cos(2\pi z_i))+e, \mathbf{z}=\mathbf{M}\cdot\mathbf{x}$ | Rotated Ackley | [-32, 32] |
| | $f_{18}(x)=\frac{1}{4000}\sum_{i=1}^{n}z_i^2+1-\prod_{i=1}^{n}\cos(\frac{z_i}{\sqrt{i}}), \mathbf{z}=\mathbf{M}\cdot\mathbf{x}$ | Rotated Griewank | [-600, 600] |

To realize comprehensive analyses and comparisons between ANS and the comparative PBSAs listed above, we conducted a series of experiments by employing 18 classical numerical optimization functions with different characteristics, including unimodality, multi-modality, rotation and noise, etc [55, 69]. The information, including type, function expression, name and search range, about each benchmark function is summarized in Table 2. In addition, to compare the efficiencies of ANS and other PBSAs in solving optimization problems with different dimensions, computational results of test functions with 30 and 100 dimensions are given. The allowed maximum function evaluations are set to 3e+5 and 6e+5 for solving functions with 30 and 100 dimensions, respectively.

With regard to ANS, the size of the population and the cardinality of superior solution collection are set to $c = m = 20$. The standard deviation of Gaussian distribution is set to $\sigma = 0.5$. Experimental analyses show that the required values of $c$ (also $m$) and $\sigma$ are relatively stable for various problems. On the contrary, the most appropriate value of $n$ may vary significantly when dealing with different problems. As currently ANS is not combined with a parameter adaptation mechanism, across-search degree $n$ is set to different values for different problems. Details about the setting of parameter $n$ are given in Table 3. Related parameter sensitivity analyses and general parameter configuration guidance are given in Section 5.

Table 3. Values of parameter $n$ for solving different optimization problems with different dimensions.

| Values | $f_1$ | $f_2$ | $f_3$ | $f_4$ | $f_5$ | $f_6$ | $f_7$ | $f_8$ | $f_9$ | $f_{10}$ | $f_{11}$ | $f_{12}$ | $f_{13}$ | $f_{14}$ | $f_{15}$ | $f_{16}$ | $f_{17}$ | $f_{18}$ |
|---|---|---|---|---|---|---|---|---|---|---|---|---|---|---|---|---|---|---|
| 30-D | 28 | 1 | 10 | 28 | 1 | 28 | 1 | 1 | 28 | 1 | 1 | 1 | 28 | 28 | 28 | 1 | 28 | 28 |
| 100-D | 8 | 20 | 8 | 10 | 10 | 10 | 1 | 1 | 10 | 1 | 1 | 1 | 10 | 10 | 8 | 1 | 20 | 40 |

The computational results for test functions with 30 dimensions are given in Table 4-5, and results of the test functions with 100 dimensions are listed in Table 6-7. In each table, "Mean" denotes the mean results obtained by running each algorithm 25 times to solve each benchmark function. "Std" stands for the related standard deviation value of the results. "NFE" denoted the average number of function evaluations required by each PBSA to obtain a satisfactory solution for each function. A run is considered to be successful if at least one solution was discovered whose fitness value is not worse than $f_i(x^*)+1e-5$, where $x^*$ denotes the global best solution [46, 55]. That is to say, if an obtained solution $x'$ which satisfy $f_i(x') \le f_i(x^*)+1e-5$, $x'$ is considered as satisfactory. The optimal objective function values of all the benchmark functions equal to zero. Therefore, in this study, we define that for any benchmark function $f_i$, if $f_i(x') < 1e-5$, $x'$ is a satisfactory solution of $f_i$. "SR" is the success rate that PBSA successfully finds a satisfactory solution for a benchmark function. "Rank" records the performance-rank of PBSAs for dealing with each benchmark function according to their obtained mean results ("Mean"). In Table 5 and 7, the rows of "Overall rank" and "Mean rank" record the overall ranks and the mean ranks of PBSAs in solving all the functions with 30 and 100 dimensions, respectively.

From the results of 30-dimension optimization functions displayed in Table 4 and 5, we can obtain some observations.

1) ANS shows good efficiency in dealing with all multimodal functions $f_7 - f_{12}$. It obtains the satisfactory solution of each multimodal function at every run. Compared with the other PBSAs, ANS ranks the best in dealing with all the benchmark multimodal problems. In addition, we can find that JADE, jDE and CLPSO also show competitive performance for multimodal problems, except that CLPSO ranks the fourth on $f_{10}$ and the third on $f_9$ while JADE ranks the fourth on $f_9$.

2) For the unimodal benchmark functions, the overall performance of ANS is also satisfactory. For example, ANS obtains the best results for functions $f_1$, $f_3$, $f_4$ and $f_5$. However, for function $f_2$, ANS performs worse than jDE and JADE. While in contrast with CLPSO, FIPS, FDR, CPSO and ABC, ANS stably shows higher efficiency.

3) The overall performance of ANS in dealing with rotated benchmark functions $f_{13} - f_{18}$ is competitive as well. For example, ANS ranks the best for dealing with functions $f_{13}$, $f_{15}$, $f_{17}$ and $f_{18}$. For functions $f_{14}$, ANS performs worse than jDE and JADE. In addition, ANS is inferior to JADE, jDE, CLPSO and FDR in solving function $f_{16}$.

4) ANS can generate the satisfactory solutions for all functions except for $f_2$, $f_6$, $f_{14}$ and $f_{16}$. The average numbers of function evaluations required for producing the satisfactory solutions are recorded in "NFE" rows, which demonstrate the good convergence of ANS. For functions $f_1$, $f_3$, $f_4$, $f_7 - f_{10}$, $f_{13}$, $f_{15}$ and $f_{17}$, ANS averagely needs the least function evaluations to obtain satisfactory solutions. For other functions, ANS generally consumes the second or third least function evaluations to successfully generate a satisfactory solution. In addition, we can find that jDE and JADE also show fairly good convergence performance. In contrast, though CLPSO can produce satisfactory solutions for many multimodal functions, its convergence is less competitive.

5) For each test function, the performance rank of each comparative PBSA is recorded. In addition, in the last two rows of Table 5, the overall rank and mean rank of each comparative PBSA on all test functions are calculated.

The overall rank and mean rank are aimed to give statistical comparison among the comparative PBSAs. We can find from Table 5 that ANS, jDE and JADE are better than CLPSO, FIPS, FDR and CPSO. Moreover, ANS is the best and slightly it is better than JADE and jDE, which rank the second and third place respectively.

It is known that the dimensionality of an optimization function exerts impacts on the efficiency of used PBSAs, i.e. the dimensionality increase generally degrades the performance of PBSAs. The computational results of each function with 100 dimensions obtained by each comparative PBSA are displayed in Table 6 and 7, from which we can also get some interesting observations.

1) Similar to the case of functions with 30 dimensions, ANS produces the best results for all the 100-dimension multimodal functions except for function $f_8$, on which ANS ranks the third.

2) For the unimodal benchmark functions, ANS produces the best results for functions $f_1$ and $f_3 - f_6$. However, it takes the fifth rank in solving function $f_2$, being inferior to jDE, JADE, FDR and CPSO.

3) It is noticeable that ANS exhibits the best performance in tackling the rotated functions with the fact that ANS produces the best results for functions $f_{13} - f_{15}$ and $f_{17} - f_{18}$. In addition, for function $f16$, it also takes the second rank.

4) The data recorded in each row of "SR" show that for function $f_{15}$, only ANS can occasionally generate satisfactory solutions. The average numbers of function evaluations required by ANS to produce the satisfactory solutions demonstrate the good convergence rate of ANS in solving the 100-dimension benchmark functions.

5) The overall rank and mean rank (given in Table 7) of each PBSA in solving all benchmark functions demonstrate that the overall performance of ANS is superior to all other comparative PBSAs. In addition, it can be observed that ANS performs better than other PBSAs on more benchmark functions when the dimensionality increases from 30 to 100. This indicates the potential of ANS in addressing high-dimension optimization problems.

From all the observations and analyses above, we can safely draw some conclusions. ANS generally owns higher efficiency in dealing with multimodal optimization problems compared to other selected comparative algorithms. Moreover, its performance in solving unimodal optimization problems is also satisfactory. For all the multimodal and unimodal problems, ANS exhibits good convergence, which indicates that ANS should have well-balanced exploitation and exploration capabilities.

Table 4. Experimental results of benchmark functions $f_1$ - $f_{11}$ with 30 dimensions. The best results are highlighted.

| Functions | | ANS | jDE | JADE | CLPSO | FIPS | FDR | CPSO | ABC |
|---|---|---|---|---|---|---|---|---|---|
| $f_1$ | Mean | **2.21E-245** | 6.32E-125 | 3.88E-131 | 6.68e-38 | 3.41e-27 | 1.38e-136 | 4.20e-12 | 6.63E-16 |
| | Std | **3.13E-244** | 1.43E-124 | 1.22E-131 | 5.41e-38 | 2.78e-27 | 1.38e-137 | 6.15e-12 | 8.12E-17 |
| | NFE | **12480** | 23070 | 22150 | 112673 | 88920 | 105301 | 97092 | 13241 |
| | SR | **100%** | 100% | 100% | 100% | 100% | 100% | 100% | 100% |
| | Rank | **1** | 4 | 3 | 5 | 6 | 2 | 8 | 7 |
| $f_2$ | Mean | 8.43E+00 | **4.86E-05** | 3.98E-01 | 1.95e+01 | 2.21e+01 | 2.25e+01 | 3.52e+01 | 8.14E+00 |
| | Std | 9.22E+00 | **8.01E-05** | 1.26E+00 | 4.31e+00 | 4.31e-01 | 3.26e+00 | 7.29e+00 | 3.22E+00 |
| | NFE | --- | **192083** | 132588 | --- | --- | --- | --- | --- |
| | SR | 0 | **30%** | 60% | 0 | 0 | 0 | 0 | 0 |
| | Rank | 3 | **1** | 2 | 5 | 7 | 6 | 8 | 4 |
| $f_3$ | Mean | **5.36E-20** | 4.15E-09 | 9.02-15 | 8.83e-03 | 3.58e-06 | 9.54e-06 | 4.10e-05 | 3.59E+00 |
| | Std | **6.44E-21** | 8.25E-10 | 1.85-14 | 2.70e-03 | 1.60e-06 | 1.10e-05 | 1.17e-04 | 5.12E-01 |
| | NFE | **88640** | 123555 | 103820 | --- | 272379 | 250311 | 241428 | --- |
| | SR | **100%** | 100% | 100% | 0 | 100% | 70% | 90% | 0 |
| | Rank | **1** | 3 | 2 | 7 | 4 | 5 | 6 | 8 |
| $f_4$ | Mean | **7.91E-168** | 2.35E-63 | 6.16E-49 | 1.89e-23 | 4.91e-17 | 2.03e-70 | 1.22e-07 | 1.49E-15 |
| | Std | **8.22E-167** | 1.35E-63 | 1.95E-48 | 9.30e-24 | 1.53e-17 | 5.80e-70 | 6.56e-08 | 1.68E-16 |
| | NFE | **11780** | 28310 | 35610 | 125302 | 108237 | 108485 | 186558 | 17980 |
| | SR | **100%** | 100% | 100% | 100% | 100% | 100% | 100% | 100% |
| | Rank | **1** | 3 | 4 | 5 | 6 | 2 | 8 | 7 |
| $f_5$ | Mean | **0.00** | **0.00** | **0.00** | **0.00** | **0.00** | **0.00** | **0.00** | **0.00** |
| | Std | **0.00** | **0.00** | **0.00** | **0.00** | **0.00** | **0.00** | **0.00** | **0.00** |
| | NFE | **9140** | **11480** | **11070** | **74184** | **39327** | **84769** | **8184** | **11253** |
| | SR | **100%** | **100%** | **100%** | **100%** | **100%** | **100%** | **100%** | **100%** |
| | Rank | **1** | **1** | **1** | **1** | **1** | **1** | **1** | **1** |
| $f_6$ | Mean | 1.54E-03 | 2.45E-03 | **6.49E-04** | 3.36e-03 | 2.62e-03 | 2.68e-03 | 9.26e-03 | 1.54E-01 |
| | Std | 5.23E-04 | 8.12E-04 | **3.31E-04** | 6.42e-04 | 1.18e-03 | 1.13e-03 | 3.64e-03 | 3.23E-02 |
| | NFE | --- | --- | **---** | --- | --- | --- | --- | --- |
| | SR | 0 | 0 | **0** | 0 | 0 | 0 | 0 | 0 |
| | Rank | 2 | 3 | **1** | 6 | 4 | 5 | 7 | 8 |
| $f_7$ | Mean | **0.00** | **0.00** | **0.00** | **0.00** | 6.50e+01 | 1.85e+01 | 1.08e-08 | 1.24E-15 |
| | Std | **0.00** | **0.00** | **0.00** | **0.00** | 1.60e+02 | 1.19e+01 | 1.42e-08 | 1.21E-15 |
| | NFE | **46500** | **53265** | **107460** | **191916** | --- | --- | 86118 | 56124 |
| | SR | **100%** | **100%** | **100%** | **100%** | 0 | 0 | 100% | 100% |
| | Rank | **1** | **1** | **1** | **1** | 8 | 7 | 6 | 5 |
| $f_8$ | Mean | **0.00** | **0.00** | **0.00** | **0.00** | 6.50e+01 | 1.85e+01 | 1.08e-08 | 2.51E-14 |
| | Std | **0.00** | **0.00** | **0.00** | **0.00** | 1.60e+02 | 1.19e+01 | 1.42e-08 | 1.38E-14 |
| | NFE | **55540** | **73370** | **127720** | **224037** | --- | --- | 207018 | 68225 |
| | SR | **100%** | **100%** | **100%** | **100%** | 0 | 0 | 100% | 100% |
| | Rank | **1** | **1** | **1** | **1** | 8 | 7 | 6 | 5 |
| $f_9$ | Mean | **3.55E-15** | **3.55E-15** | 4.44E-15 | 4.26e-15 | 1.74e-14 | 2.59e-14 | 3.93e-07 | 5.15E-14 |
| | Std | **0.00** | **0.00** | 0.00 | 1.45e-15 | 5.90e-15 | 1.34e-14 | 4.76e-07 | 5.25E-15 |
| | NFE | **15300** | **32255** | 31530 | 137809 | 124512 | 117742 | 219852 | 23542 |
| | SR | **100%** | **100%** | 100% | 100% | 100% | 100% | 100% | 100% |
| | Rank | **1** | **1** | 4 | 3 | 5 | 6 | 8 | 7 |
| $f_{10}$ | Mean | **0.00** | **0.00** | **0.00** | 9.99e-17 | 4.19e-08 | 1.25e-02 | 1.88e-02 | 4.10E-14 |
| | Std | **0.00** | **0.00** | **0.00** | 3.15e-16 | 1.24e-07 | 7.9e-03 | 2.51e-02 | 2.44E-14 |
| | NFE | **19340** | **24905** | **24070** | 138211 | 129954 | 104151 | 109256 | 19778 |
| | SR | **100%** | **100%** | **100%** | 100% | 100% | 10% | 30% | 100% |
| | Rank | **1** | **1** | **1** | 4 | 6 | 7 | 8 | 5 |

Table 5. Experimental results of benchmark functions $f_{12}$ - $f_{18}$ with 30 dimensions. The best results are highlighted.

| Functions | | ANS | jDE | JADE | CLPSO | FIPS | FDR | CPSO | ABC |
|---|---|---|---|---|---|---|---|---|---|
| $f_{11}$ | Mean | **1.57E-32** | **1.57E-32** | **1.57E-32** | **1.57E-32** | 2.44e-27 | **1.57E-32** | 1.17e-14 | 5.96E-16 |
| | Std | **2.72E-48** | **2.88E-48** | **2.88E-48** | **2.88E-48** | 2.83e-27 | **2.88E-48** | 1.65e-14 | 5.65E-16 |
| | NFE | **15420** | **23625** | **20900** | **102606** | 84939 | **96735** | 45012 | 14547 |
| | SR | **100%** | **100%** | **100%** | **100%** | 100% | **100%** | 100% | 100% |
| | Rank | **1** | **1** | **1** | **1** | 7 | **1** | 6 | 8 |
| $f_{12}$ | Mean | **1.35E-32** | **1.35E-32** | **1.35E-32** | **1.35E-32** | 1.25e-27 | 5.49e-03 | 4.88e-13 | 6.66E-16 |
| | Std | **2.88E-32** | **2.88E-32** | **2.88E-32** | **2.88E-32** | 6.42e-28 | 5.79e-03 | 1.26e-14 | 8.45E-17 |
| | NFE | **17080** | **23715** | **22330** | **105631** | 84933 | 101699 | 60915 | 15487 |
| | SR | **100%** | **100%** | **100%** | **100%** | 100% | 50% | 100% | 100% |
| | Rank | **1** | **1** | **1** | **1** | 5 | 8 | 7 | 6 |
| $f_{13}$ | Mean | **1.71E-199** | 1.39E-104 | 1.38E-119 | 2.80e-17 | 4.71e-24 | 7.01e-131 | 6.82e-12 | 1.79E-15 |
| | Std | **1.35E-201** | 2.45E-104 | 4.32E-119 | 3.77e-17 | 4.44e-24 | 1.49e-130 | 8.04e-12 | 5.44E-16 |
| | NFE | **16320** | 28135 | 27850 | 143677 | 100401 | 110623 | 117738 | 31525 |
| | SR | **100%** | 100% | 100% | 100% | 100% | 100% | 100% | 100% |
| | Rank | **1** | 4 | 3 | 6 | 5 | 2 | 8 | 7 |
| $f_{14}$ | Mean | 1.82E+01 | **1.73E-01** | 4.05E-01 | 2.62e+01 | 2.69e+01 | 2.18e+01 | 2.59e+01 | 2.44E+01 |
| | Std | 6.32E+00 | **3.19E+00** | 1.28E+00 | 9.10e-01 | 1.22e+00 | 3.13e+00 | 1.99e+01 | 1.85E+00 |
| | NFE | --- | --- | 294800 | --- | --- | --- | --- | --- |
| | SR | 0 | **0** | 20 | 0 | 0 | 0 | 0 | 0 |
| | Rank | 3 | **1** | 2 | 7 | 8 | 4 | 6 | 5 |
| $f_{15}$ | Mean | **1.32E-45** | 2.12e-21 | 1.62e-31 | 1.36e-02 | 2.33e-09 | 1.99e-02 | 2.75e+00 | 3.27E+00 |
| | Std | **2.52e-11** | 4.42e-22 | 2.32e-32 | 2.78e-03 | 1.02e-09 | 3.09e-02 | 5.07e-01 | 3.54E-01 |
| | NFE | **39300** | 68715 | 51890 | --- | 176985 | --- | --- | --- |
| | SR | **100%** | 100% | 100% | 0 | 100% | 0 | 0 | 0 |
| | Rank | **1** | 3 | 2 | 5 | 4 | 6 | 7 | 8 |
| $f_{16}$ | Mean | 1.61E+02 | 3.75E+01 | **2.16E+01** | 1.15e+01 | 1.77e+02 | 5.12e+01 | 3.66e+02 | 2.90E+02 |
| | Std | 3.15E+01 | 8.69E+00 | **4.08E+00** | 1.50e+01 | 9.85e+00 | 1.35e+01 | 4.91e+01 | 3.42E+01 |
| | NFE | --- | --- | --- | --- | --- | --- | --- | --- |
| | SR | 0 | 0 | **0** | 0 | 0 | 0 | 0 | |
| | Rank | 5 | 2 | **1** | 4 | 6 | 3 | 8 | 7 |
| $f_{17}$ | Mean | **3.55E-15** | 6.35E-14 | 7.55E-14 | 3.48e-03 | 2.16e-14 | 1.73e+00 | 1.67e+01 | 1.12E+01 |
| | Std | **4.98E-16** | 5.21E-15 | 6.31E-15 | 4.66e-03 | 7.19e-15 | 5.24e+01 | 4.98e+00 | 3.85E+00 |
| | NFE | **16700** | 33895 | 31580 | --- | 128283 | --- | --- | --- |
| | SR | **100%** | 100% | 100% | 0 | 100% | 0 | 0 | 0 |
| | Rank | **1** | 2 | 3 | 5 | 4 | 6 | 8 | 7 |
| $f_{18}$ | Mean | **4.62E-16** | 5.66E-03 | 4.44E-03 | 3.05e-03 | 3.43e-03 | 1.33e-02 | 1.57e-02 | 4.42E-04 |
| | Std | **8.73E-17** | 6.54E-03 | 6.23E-03 | 4.80e-03 | 6.92e-03 | 1.31e-02 | 1.31e-02 | 7.54E-04 |
| | NFE | **98648** | 77630 | 68283 | --- | 200757 | 128398 | --- | 111258 |
| | SR | **100%** | 50% | 60% | 0 | 40% | 20% | 0 | 0 |
| | Rank | **1** | 4 | 2 | 5 | 6 | 7 | 8 | 2 |
| Mean rank | | **1.50** | 2.06 | 1.94 | 4.00 | 5.56 | 4.72 | 6.89 | 5.94 |
| Overall rank | | **1** | 3 | 2 | 4 | 6 | 5 | 8 | **7** |

Table 6. Experimental results of benchmark functions $f_1$ - $f_{11}$ with 100 dimensions. The best results are highlighted.

| Functions | | ANS | jDE | JADE | CLPSO | FIPS | FDR | CPSO | ABC |
|---|---|---|---|---|---|---|---|---|---|
| $f_1$ | Mean | **8.40E-134** | 8.25E-73 | 1.54E-35 | 3.85E-23 | 8.41E-05 | 1.78E-41 | 1.43e-07 | 5.37E-15 |
| | Std | **7.56E-134** | 6.21E-72 | 1.45E-35 | 1.77E-23 | 2.47E-05 | 4.76E-41 | 2.56e-07 | 6.19E-16 |
| | SR | **49560** | 81810 | 37290 | 291884 | --- | 295613 | 369357 | 50248 |
| | NFE | **100%** | 100% | 100% | 100% | 0 | 100% | 100% | 100% |
| | Rank | **1** | 2 | 4 | 5 | 8 | 3 | 7 | 6 |
| $f_2$ | Mean | 8.28E+01 | **6.11E+01** | 7.96E+01 | 8.91E+01 | 9.50E+01 | 6.52E+01 | 8.01e+01 | 9.63E+01 |
| | Std | 7.18E+00 | **6.97e+00** | 1.62E+01 | 6.67E+00 | 6.51E-01 | 5.78E+00 | 3.96e+01 | 2.22E+01 |
| | SR | --- | **---** | --- | --- | --- | --- | --- | --- |
| | NFE | 0 | **0** | 0 | 0 | 0 | 0 | 0 | 0 |
| | Rank | 5 | **1** | 3 | 6 | 8 | 2 | 4 | 7 |
| $f_3$ | Mean | **9.82E-05** | 1.47E-01 | 1.06E+00 | 2.12E-01 | 8.91E-01 | 6.36E-01 | 1.76e+00 | 8.64E+00 |
| | Std | **2.54E-04** | 2.55E-01 | 1.47E-01 | 3.47E-02 | 4.24E-02 | 1.01E-01 | 2.76e-01 | 2.49E-01 |
| | SR | **---** | --- | --- | --- | --- | --- | --- | -- |
| | NFE | **0** | 0 | 0 | 0 | 0 | 0 | 0 | 0 |
| | Rank | **1** | 2 | 6 | 3 | 5 | 4 | 7 | 8 |
| $f_4$ | Mean | **1.47E-90** | 2.71E-66 | 1.07E-20 | 3.85E-15 | 7.49E-05 | 8.51E-22 | 3.99e-05 | 9.36E-15 |
| | Std | **1.24E-90** | 3.82E-67 | 1.03E-20 | 6.54E-16 | 1.55E-05 | 1.61E-21 | 1.26e-05 | 8.75E-16 |
| | SR | **48380** | 120254 | 51900 | 320266 | --- | 309088 | --- | 68680 |
| | NFE | **100%** | 100% | 100% | 100% | 0 | 100% | 0 | |
| | Rank | **1** | 2 | 4 | 5 | 8 | 3 | 7 | 6 |
| $f_5$ | Mean | **0.00** | **0.00** | **0.00** | **0.00** | **0.00** | 3.98e+01 | **0.00** | 2.50E+00 |
| | Std | **0.00** | **0.00** | **0.00** | **0.00** | **0.00** | 1.64e+01 | **0.00** | 1.17E+00 |
| | SR | **23540** | **44060** | **38350** | **189490** | **318522** | --- | **33633** | --- |
| | NFE | **100%** | **100%** | **100%** | **100%** | **100%** | 0 | **100%** | 0 |
| | Rank | **1** | **1** | **1** | **1** | **1** | 8 | **1** | 7 |
| $f_6$ | Mean | **1.28e-02** | 3.54e-02 | 3.22e-02 | 3.12e-02 | 8.47e-02 | 6.50e-02 | 3.99e-02 | 1.51E+00 |
| | Std | **8.44e-03** | 2.67e-02 | 9.02e-04 | 2.42e-03 | 7.28e-03 | 1.55e-02 | 9.17e-03 | 2.59E-01 |
| | SR | **---** | --- | --- | --- | --- | --- | --- | --- |
| | NFE | **0** | 0 | 0 | 0 | 0 | 0 | 0 | 0 |
| | Rank | **1** | 4 | 3 | 2 | 7 | 6 | 5 | 8 |
| $f_7$ | Mean | **0.00** | **0.00** | 7.25E+01 | 8.91E-12 | 6.54E+02 | 1.80E+02 | 4.51e-08 | 3.57E-11 |
| | Std | **0.00** | **0.00** | 2.16E+00 | 8.34E-12 | 2.81E+01 | 2.98E+01 | 5.95e-08 | 8.81E-11 |
| | SR | **158680** | **154130** | --- | 491601 | --- | --- | 341784 | 342150 |
| | NFE | **100%** | **100%** | 0 | 100% | 0 | 0 | 100% | 100% |
| | Rank | **1** | **1** | 6 | 3 | 8 | 7 | 5 | 4 |
| $f_8$ | Mean | 2.66E-01 | 1.59E+01 | 7.23E+01 | 6.00E-01 | 8.57E+02 | 1.98E+02 | 2.02e-04 | **1.38E-06** |
| | Std | 0.00 | 4.54E+01 | 2.13E+00 | 8.43E-01 | 3.62E+01 | 1.43E+02 | 2.04e-05 | **4.06E-06** |
| | SR | 188660 | 528975 | --- | 587084 | --- | --- | 355956 | **432412** |
| | NFE | 75% | 40% | 0 | 60% | 0 | 0 | 60% | **80%** |
| | Rank | 3 | 5 | 6 | 4 | 8 | 7 | 2 | **1** |
| $f_9$ | Mean | **2.62E-14** | 8.79E-01 | 1.43E-14 | 6.54E-13 | 1.03E-03 | 2.83E-01 | 3.74e-05 | 3.76E-13 |
| | Std | **1.52E-15** | 2.78E-01 | 1.49E-15 | 1.41E-13 | 2.05E-04 | 6.01E-01 | 2.44e-05 | 7.04E-14 |
| | SR | **57260** | 88872 | 49340 | 346480 | --- | 340855 | --- | 80558 |
| | NFE | **100%** | 90% | 100% | 100% | 0 | 80% | 0 | 100% |
| | Rank | **1** | 8 | 2 | 4 | 6 | 7 | 5 | 3 |
| $f_{10}$ | Mean | **0.00** | 5.85E-03 | 1.72E-03 | **0.00** | 4.24E-04 | 2.95E-03 | 7.14e-03 | 4.09E-14 |
| | Std | **0.00** | 1.85E-02 | 3.68E-03 | **0.00** | 6.60E-04 | 9.33E-03 | 8.94e-03 | 6.81E-14 |
| | SR | **59820** | 82327 | 36962 | **296716** | --- | 292062 | 375114 | 62365 |
| | NFE | **100%** | 90% | 80% | **100%** | 0 | 90% | 50% | 100% |
| | Rank | **1** | 7 | 5 | **1** | 4 | 6 | 8 | 3 |

Table 7. Experimental results of benchmark functions $f_{12}$ - $f_{18}$ with 100 dimensions. The best results are highlighted.

| Functions | | ANS | jDE | JADE | CLPSO | FIPS | FDR | CPSO | ABC |
|---|---|---|---|---|---|---|---|---|---|
| $f_{11}$ | Mean | **1.57E-32** | **1.57E-32** | **1.57E-32** | 1.55E-24 | 7.40E+00 | 2.81E-01 | 9.58e-11 | 5.76E-15 |
| | Std | **2.88E-32** | **2.88E-32** | **2.88E-32** | 6.43E-25 | 5.95E+00 | 7.14E-01 | 5.20e-11 | 1.32E-15 |
| | SR | **54640** | **50295** | **28490** | 267106 | --- | 284219 | 159984 | 51875 |
| | NFE | **100%** | **100%** | **100%** | 100% | 0 | 60% | 100% | 100% |
| | Rank | **1** | **1** | **1** | 4 | 8 | 7 | 6 | 5 |
| $f_{12}$ | Mean | **1.35E-32** | 2.51e-32 | 6.37E-32 | 1.53E-24 | 1.32E-03 | 1.75E-02 | 1.59e-09 | 6.42E-15 |
| | Std | **2.88E-32** | 3.52e-32 | 3.42E-33 | 6.94E-25 | 6.98E-04 | 3.19E-02 | 1.26e-09 | 2.64E-15 |
| | SR | **54360** | 51810 | 55600 | 267622 | --- | 272210 | 241188 | 47230 |
| | NFE | **100%** | 100% | 100% | 100% | 0 | 60% | 100% | 100% |
| | Rank | **1** | 2 | 3 | 4 | 7 | 8 | 6 | 5 |
| $f_{13}$ | Mean | **8.01E-133** | 3.11e-85 | 9.41e-36 | 3.83e-23 | 1.06e-04 | 1.71e-42 | 7.08e-08 | 4.94E-15 |
| | Std | **5.32E-134** | 3.76e-85 | 6.17e-36 | 1.21e-23 | 2.84e-05 | 5.18e-42 | 6.14e-08 | 1.11E-15 |
| | SR | **49380** | 58295 | 67010 | 292361 | --- | 295620 | 397233 | 52320 |
| | NFE | **100%** | 100% | 100% | 100% | 0 | 100% | 100% | 100% |
| | Rank | **1** | 2 | 4 | 5 | 8 | 3 | 7 | 6 |
| $f_{14}$ | Mean | **8.42E+01** | 8.93e+01 | 9.83e+01 | 9.16e+01 | 1.02e+02 | 9.00e+01 | 1.50e+02 | 9.13E+01 |
| | Std | **3.68E+00** | 1.92e+01 | 6.88e-01 | 3.79e+00 | 1.84e+01 | 1.61e+01 | 3.28e+01 | 4.30E+00 |
| | SR | **---** | --- | --- | --- | --- | --- | --- | --- |
| | NFE | **0** | 0 | 0 | 0 | 0 | 0 | 0 | 0 |
| | Rank | **1** | 2 | 6 | 5 | 7 | 3 | 8 | 4 |
| $f_{15}$ | Mean | **3.73E-05** | 3.99e-03 | 1.43e-01 | 1.24e-01 | 7.36e-01 | 4.28e-01 | 2.85e+00 | 7.81E+00 |
| | Std | **3.44E-05** | 9.70e-03 | 4.18e-02 | 3.15e-02 | 6.59e-02 | 1.12e-01 | 5.98e-01 | 4.52E-01 |
| | SR | **228425** | --- | --- | --- | --- | --- | --- | -- |
| | NFE | **20%** | 0 | 0 | 0 | 0 | 0 | 0 | 0 |
| | Rank | **1** | 2 | 4 | 3 | 6 | 5 | 7 | 8 |
| $f_{16}$ | Mean | 2.17e+02 | **1.35e+02** | 2.65e+02 | 2.29e+02 | 8.02e+02 | 2.26e+02 | 3.29e+02 | 3.18E+02 |
| | Std | 2.22e+01 | **1.51e+01** | 1.11e+01 | 2.73e+01 | 3.36e+01 | 2.22e+01 | 5.56e+01 | 1.67E+01 |
| | SR | --- | **---** | --- | --- | --- | --- | --- | --- |
| | NFE | 0 | **0** | 0 | 0 | 0 | 0 | 0 | 0 |
| | Rank | 2 | **1** | 5 | 4 | 8 | 3 | 6 | 7 |
| $f_{17}$ | Mean | **1.12e-14** | 1.09e-01 | 1.20e-14 | 6.57e-07 | 1.74e-03 | 2.77e+00 | 2.04e+00 | 1.90E+00 |
| | Std | **0.00** | 3.45e-01 | 2.99e-15 | 9.73e-07 | 2.84e-04 | 5.38e-01 | 3.63e-01 | 1.49E-01 |
| | SR | **67135** | 86422 | 69840 | 502077 | --- | --- | --- | --- |
| | NFE | **100%** | 90% | 100% | 100% | 0 | 0 | 0 | 0 |
| | Rank | **1** | 5 | 2 | 3 | 4 | 8 | 7 | 6 |
| $f_{18}$ | Mean | **7.41e-18** | 4.13e-02 | 2.70e-03 | 4.91e-08 | 1.49e-03 | 2.96e-03 | 2.26e-02 | 3.68E-07 |
| | Std | **1.22e-17** | 1.17e01 | 5.74e-03 | 2.53e-08 | 1.42e-03 | 6.66-03 | 2.29e-02 | 5.58E-07 |
| | SR | **76060** | 127640 | 71737 | 405472 | --- | 309347 | 461570 | 145480 |
| | NFE | **100%** | 70% | 80% | 100% | 0 | 80% | 30% | 100% |
| | Rank | **1** | 8 | 5 | 2 | 4 | 6 | 7 | 3 |
| Mean rank | | **1.39** | 3.11 | 3.89 | 3.56 | 6.39 | 5.33 | 5.83 | 5.39 |
| Overall rank | | **1** | 2 | 4 | 3 | 8 | 5 | 7 | 6 |

### 4.2 Statistical analyses of the results

As all the comparative EAs and SIAs in experiments are stochastic, the results obtained by each algorithm exhibit randomness. In experiments, each algorithm is run 25 times when solving each benchmark function. To realize reliable comparisons, significance tests are necessary. Recently, nonparametric statistical test is thought to be more appropriate in evolutionary computation area for several advantages [9, 51]. Hence, in this paper, the nonparametric Wilcoxon's rank sum test at a 0.05 significance level is conducted between ANS and each peer algorithm on their respective 25 results produced by running each algorithm 25 times. "-", "+", and "≈" indicate that the performance of the corresponding peer algorithm is worse than, better than, and similar to that of ANS, respectively. Results of Wilcoxon's rank sum test over functions with 30 and 100 dimensions are reported in Table 8 and Table 9, respectively.

As for functions with 30 dimensions, it can be observed from Table 8 that, ANS is better than jDE, JADE, CLPSO, FIPS, FDR, CPSO and ABC on 8, 8, 13, 17, 16, 17 and 17 functions while similar to them on 7, 6, 5, 1, 2, 1 and 1 function, respectively. ANS is inferior to jDE and JADE on 3 and 4 functions, respectively. It is worth noting that CLPSO, FIPS, FDR, CPSO and ABC cannot beat ANS on any function.

With regard to functions with 100 dimensions, we can find from Table 9 that, ANS is superior to jDE, JADE, CLPSO, FIPS, FDR, CPSO and ABC on 13, 15, 16, 17, 17, 15 and 17 functions, similar to them on 3, 2, 2, 1, 0, 1, and 0 function, and inferior to them on 2, 1, 0, 0, 1, 2 and 1 function, respectively.

Table 8. The results of Wilcoxon's rank sum test over each function with 30 dimensions

| Results | vs jDE | vs JADE | vs CLPSO | vs FIPS | vs FDR | vs CPSO | vs ABC |
|---|---|---|---|---|---|---|---|
| $f_1$ | — | — | — | — | — | — | — |
| $f_2$ | + | + | — | — | — | — | — |
| $f_3$ | — | — | — | — | — | — | — |
| $f_4$ | — | — | — | — | — | — | — |
| $f_5$ | ≈ | ≈ | ≈ | ≈ | ≈ | ≈ | ≈ |
| $f_6$ | — | + | — | — | — | — | — |
| $f_7$ | ≈ | ≈ | ≈ | — | — | — | — |
| $f_8$ | ≈ | ≈ | ≈ | — | — | — | — |
| $f_9$ | ≈ | — | — | — | — | — | — |
| $f_{10}$ | ≈ | ≈ | — | — | — | — | — |
| $f_{11}$ | ≈ | ≈ | ≈ | — | ≈ | — | — |
| $f_{12}$ | ≈ | ≈ | ≈ | — | — | — | — |
| $f_{13}$ | — | — | — | — | — | — | — |
| $f_{14}$ | + | + | — | — | — | — | — |
| $f_{15}$ | — | — | — | — | — | — | — |
| $f_{16}$ | + | + | + | — | — | — | — |
| $f_{17}$ | — | — | — | — | — | — | — |
| $f_{18}$ | — | — | — | — | — | — | — |
| — | 8 | 8 | 13 | 17 | 16 | 17 | 17 |
| + | 3 | 4 | 0 | 0 | 0 | 0 | 0 |
| ≈ | 7 | 6 | 5 | 1 | 2 | 1 | 1 |

Table 9. The results of Wilcoxon's rank sum test over each function with 100 dimensions

| Results | vs jDE | vs JADE | vs CLPSO | vs FIPS | vs FDR | vs CPSO | vs ABC |
|---|---|---|---|---|---|---|---|
| $f_1$ | — | — | — | — | — | — | — |
| $f_2$ | + | + | — | — | + | + | — |
| $f_3$ | — | — | — | — | — | — | — |
| $f_4$ | — | — | — | — | — | — | — |
| $f_5$ | ≈ | ≈ | ≈ | ≈ | — | ≈ | — |
| $f_6$ | — | — | — | — | — | — | — |
| $f_7$ | ≈ | — | — | — | — | — | — |
| $f_8$ | — | — | — | — | — | + | + |
| $f_9$ | — | — | — | — | — | — | — |
| $f_{10}$ | — | — | ≈ | — | — | — | — |
| $f_{11}$ | ≈ | ≈ | — | — | — | — | — |
| $f_{12}$ | — | — | — | — | — | — | — |
| $f_{13}$ | — | — | — | — | — | — | — |
| $f_{14}$ | — | — | — | — | — | — | — |
| $f_{15}$ | — | — | — | — | — | — | — |
| $f_{16}$ | + | — | — | — | — | — | — |
| $f_{17}$ | — | — | — | — | — | — | — |
| $f_{18}$ | — | — | — | — | — | — | — |
| — | 13 | 15 | 16 | 17 | 17 | 15 | 17 |
| + | 2 | 1 | 0 | 0 | 1 | 2 | 1 |
| ≈ | 3 | 2 | 2 | 1 | 0 | 1 | 0 |

Although the significance tests between ANS and its comparative algorithms are conducted on each problem, one may wonder whether and how the performance of ANS is significantly improved over other algorithms with respect to the whole benchmark suit. As a result, the 1x7 Wilcoxon signed-rank test is conducted between ANS and other algorithms first. Corresponding *p*-values can be obtained by the 1x7 Wilcoxon signed-rank test. However, as suggested in [9], these *p*-values are not suitable for multiple comparisons. When a *p*-value is considered in a multiple test, it reflects the probability error of a certain comparison, but it does not take into account the remaining comparisons belonging to the family. Adjusted *p*-values (APVs) can deal with this shortcoming. Therefore, Finner post-hoc analysis procedure [9, 13] is utilized to calculate the APVs. The related *p*-values and APVs for functions with 30 and 100 dimensions are reported in Table 10 and Table 11, respectively.

From the data of the benchmark functions with 30 dimensions given in Table 10, we can find that, given a 0.1 significance level, the 1x7 Wilcoxon signed-rank test confirms the improvements of ANS over FIPS, CPSO, ABC and FDR for the post-hoc procedure. In contrast, significant differences between ANS and CLPSO, JADE and jDE are not detected.

In addition, as for the benchmark functions with 100 dimensions, the *p*-values and APVs data listed in Table 11 show that, given a 0.1 significance level, the 1x7 Wilcoxon signed-rank test demonstrates the significant improvement of ANS over FIPS, CLPSO, ABC, FDR, JADE and CPSO for the post-hoc procedure. The difference between ANS and jDE is not significant yet.

Table 10. Results of the 1 x 7 Wilcoxon signed-rank test and post-hoc analysis on all functions with 30 dimensions

| **Comparison** | ***p*-values** | **Finner adjusted *p*-values** |
|---|---|---|
| ANS versus FIPS | 2.9248E-04 | 2.0456E-03 |
| ANS versus CPSO | 2.9305E-04 | 2.0496E-03 |
| ANS versus ABC | 2.9305E-04 | 2.0496E-03 |
| ANS versus FDR | 7.1601E-03 | 4.9057E-02 |
| ANS versus CLPSO | 3.5278E-02 | 2.2230E-01 |
| ANS versus JADE | 3.7573E-01 | 9.6305E-01 |
| ANS versus jDE | 8.0078E-01 | 9.9999E-01 |

Table 11. Results of the 1 x 7 Wilcoxon signed-rank test and post-hoc analysis on all functions with 100 dimensions

| **Comparison** | ***p*-values** | **Finner adjusted *p*-values** |
|---|---|---|
| ANS versus FIPS | 2.9305E-04 | 2.0496E-03 |
| ANS versus CLPSO | 4.3778E-04 | 3.0604E-03 |
| ANS versus ABC | 1.0088E-03 | 7.0405E-03 |
| ANS versus FDR | 2.1384E-03 | 1.4873E-02 |
| ANS versus JADE | 9.7255E-03 | 6.6124E-02 |
| ANS versus CPSO | 1.4772E-02 | 9.8935E-02 |
| ANS versus jDE | 8.3252E-02 | 4.5581E-01 |

**4.2. Experimental comparison between ANS and some other PBSAs**

In the previous section, ANS has been compared with some very efficient and popular PSO and DE variants. However, there are still some other conventional PBSAs and particularly some new PBSAs that have emerged in quite recent years. In this section, the performance of ANS is compared with some new emerged PBSAs and some conventional PBSAs.

The selected new emerged comparative PBSAs are listed below.

**RCCRO1:** real-coded chemical reaction optimization [32].

**RCBBO:** real-coded biogeography-based optimization [18], RCBBO is an improved version of biogeography based optimization [48] on solving problems in the continuous domain.

**GSO:** group search optimizer [20].

The conventional PBSAs selected for the comparison in this section are given as follows.

**CMAES:** covariance matrix adaptation evolution strategy [19].

**G3PCX:** generalized generation gap model with generic parent-centric recombination operator [8].

**FEP:** fast evolutionary programming [67].

**FES:** fast evolutionary strategy [66].

CMAES and G3PCX are very competitive PBSAs and they are considered as benchmark algorithms for numerical optimization [32].

The comparison results are listed in Table 13. As functions $f_1 - f_7$ and $f_9 - f_{12}$ were also taken as the test functions in [32], these functions are utilized in the comparison. The results of RCCRO1, RCBBO, GSO, CMAES, G3PCX, FEP and FES are from the Table V and VI of [32]. As suggested by Mernik et. al. [39] and Črepinšek et. al. [5], all experiments should be conducted and compared under the same conditions. In [32], different function evaluation amounts are adopted in solving different benchmark functions. To make fair comparisons, the number of function evaluations used by ANS for solving each function is consistent with those reported in Table II of [32]. The detailed function evaluation numbers used for test functions are listed in the "NFEs" column of Table 13.

From the results given in Table 13, we can discover that, compared with the selected new and conventional PBSAs, ANS produce the best results for most of the benchmark functions, except for functions $f_2$ and $f_3$. While for function $f_3$, the performance of ANS is still quite satisfactory with being only worse than CMAES. For function $f_2$, ANS exhibits medium performance.

The overall performance of each PBSA measured by the overall rank and mean rank are given in the last two rows of Table 13. The results show that, with regard to the considered test functions, ANS performs better than other comparative algorithms. This indicates that ANS could be a useful alternative algorithm for solving numerical optimization problems.

Table 13 Comparison results between ANS and some new and conventional PBSAs. The best results are highlighted.

| Function | NFEs | | ANS | RCCRO1 | GSO | RCBBO | CMAES | G3PCX | FEP | FES |
|---|---|---|---|---|---|---|---|---|---|---|
| $f_1$ | 150 000 | Mean | **4.48E-121** | 6.43E−07 | 1.95E−08 | 1.39E−03 | 6.09E−29 | 6.40E−79 | 5.70E−04 | 2.50E−04 |
| | | Std | **2.25E-120** | 2.10E−07 | 1.16E−08 | 5.50E−04 | 1.55E−29 | 1.25E−78 | 1.30E−04 | 6.80E−04 |
| | | Rank | **1** | 5 | 4 | 8 | 3 | 2 | 7 | 6 |
| $f_2$ | 150 000 | Mean | 1.77E+01 | 2.71E+01 | 4.98E+01 | 5.54E+01 | **5.58E−01** | 3.09E+00 | 5.06E+00 | 3.33E+01 |
| | | Std | 7.21E+00 | 3.43E+01 | 3.02E+01 | 3.52E+01 | **1.39E+00** | 1.64E+01 | 5.87E+00 | 4.31E+01 |
| | | Rank | 4 | 5 | 7 | 8 | **1** | 2 | 3 | 6 |
| $f_3$ | 150 000 | Mean | 1.33E-10 | 9.32E−03 | 1.08E−01 | 3.09E−02 | **3.99E−15** | 4.53E+01 | 3.00E−01 | 5.50E−03 |
| | | Std | 2.57E-11 | 3.66E−03 | 3.99E−02 | 7.27E−03 | **5.31E−16** | 8.09E+00 | 5.00E−01 | 6.50E−04 |
| | | Rank | 2 | 4 | 6 | 5 | **1** | 8 | 7 | 3 |
| $f_4$ | 150 000 | Mean | **3.39E-83** | 2.19E−03 | 3.70E−05 | 7.99E−02 | 3.48E−14 | 2.83E+01 | 8.10E−03 | 6.00E−02 |
| | | Std | **5.22E-82** | 4.34E−04 | 8.61E−05 | 1.44E−02 | 4.03E−15 | 1.01E+01 | 7.70E−04 | 9.60E−03 |
| | | Rank | **1** | 4 | 3 | 7 | 2 | 8 | 5 | 6 |
| $f_5$ | 150 000 | Mean | **0.00E+00** | **0.00E+00** | 1.60E−02 | **0.00E+00** | 7.00E−02 | 9.42E+01 | **0.00E+00** | **0.00E+00** |
| | | Std | **0.00E+00** | **0.00E+00** | 1.33E−01 | **0.00E+00** | 2.93E−01 | 5.96E+01 | **0.00E+00** | **0.00E+00** |
| | | Rank | **1** | **1** | 6 | **1** | 7 | 8 | **1** | **1** |
| $f_6$ | 150 000 | Mean | **3.05E-03** | 5.40E−03 | 7.37E−02 | 1.75E−02 | 2.21E−01 | 9.79E−01 | 7.60E−03 | 1.20E−02 |
| | | Std | **2.21E-03** | 2.98E−03 | 9.25E−02 | 6.43E−03 | 8.65E−02 | 4.63E−01 | 2.60E−03 | 5.80E−03 |
| | | Rank | **1** | 2 | 6 | 5 | 7 | 8 | 3 | 4 |
| $f_7$ | 250 000 | Mean | **0.00** | 9.08E−04 | 1.02E+00 | 2.62E−02 | 4.95E+01 | 1.74E+02 | 4.60E−02 | 1.60E−01 |
| | | Std | **0.00** | 2.88E−04 | 9.51E-01 | 9.76E-03 | 1.23E+01 | 3.19E+01 | 1.20E−02 | 3.30E−01 |
| | | Rank | **1** | 2 | 6 | 3 | 7 | 8 | 4 | 5 |
| $f_9$ | 150 000 | Mean | **5.68E-15** | 1.94E−03 | 2.66E−05 | 2.51E−02 | 4.61E+00 | 1.35E+01 | 1.80E−02 | 1.20E−02 |
| | | Std | **3.17E-16** | 4.19E−04 | 3.08E−05 | 5.51E−03 | 8.73E+00 | 4.82E+00 | 2.10E−02 | 1.80E−03 |
| | | Rank | **1** | 3 | 2 | 6 | 7 | 8 | 5 | 4 |
| $f_{10}$ | 150 000 | Mean | **0.00** | 1.12E−02 | 3.08E−02 | 4.82E−01 | 7.39E−04 | 1.12E−02 | 1.60E−02 | 3.70E−02 |
| | | Std | **0.00** | 1.62E−02 | 3.09E−02 | 8.49E−02 | 2.38E−03 | 1.31E−02 | 2.20E−02 | 5.00E−02 |
| | | Rank | **1** | 4 | 6 | 8 | 2 | 3 | 5 | 7 |
| $f_{11}$ | 150 000 | Mean | **1.57E-32** | 2.07E−02 | 2.77E−11 | 3.28E−05 | 5.17E−03 | 4.59E+00 | 9.20E−06 | 2.80E−02 |
| | | Std | **2.88E-48** | 5.49E−02 | 9.17E−11 | 3.33E−05 | 7.34E−03 | 5.98E+00 | 6.14E−05 | 8.10E−11 |
| | | Rank | **1** | 6 | 2 | 4 | 5 | 8 | 3 | 7 |
| $f_{12}$ | 150 000 | Mean | **1.35E-32** | 7.05E−07 | 4.69E−05 | 3.72E−04 | 1.64E−03 | 2.35E+01 | 1.60E−04 | 4.70E−05 |
| | | Std | **0.00** | 5.90E−07 | 7.00E−04 | 4.63E−04 | 4.19E−03 | 2.07E+01 | 7.30E−05 | 1.50E−05 |
| | | Rank | **1** | 2 | 3 | 6 | 7 | 8 | 5 | 4 |
| Mean rank | | | **1.36** | 3.45 | 4.64 | 5.55 | 4.45 | 6.45 | 4.36 | 4.82 |
| Overall rank | | | **1** | 2 | 5 | 8 | 4 | 7 | 3 | 6 |

**5. Parameter analyses**

In this section, extensive experiments are conducted to analyze the impact of each parameter on the performance of ANS. As the aim here is to provide readers a straightforward and intuitive impression on how different parameter values affect ANS, only the data that thought to be informative are presented. Generally, the appropriate value of across-search degree $n$ varies in the case of different functions or different dimensionalities, while the required values of population size $m$ and standard deviation $\sigma$ are relatively stable. Hence, the analysis data of $n$ on functions both with 30 and 100 dimensions are given, whereas the presented analysis data of $m$ and $\sigma$ are restricted to functions with 30 dimensions.

The candidate values for $n$ to be analysed are [1, 2, 4, 6, 8, 10, 12, 14, 16, 18, 20, 22, 24, 26, 28] and [1, 2, 4, 6, 8, 10, 20, 30, 40, 50, 60, 70, 80, 90, 98] for functions with 30 and 100 dimensions, respectively. The candidate values for $\sigma$ to be analysed are [0.1, 0.2, 0.3, 0.4, 0.5, 0.6, 0.7, 0.8, 0.9, 1.0], and the candidate values for $ps$ to be analysed are [5, 10, 20, 30, 40, 50].

In parameter sensitivity analysis, full factorial designs and fractional factorial designs (e.g. orthogonal experiment design) are thought to be more reliable [33]. However, they are not adopted currently in this study because the computation cost is too expensive. There are three reasons. First, the number of candidate values of each parameter is relatively high (i.e. 15 for $n$, 10 for $\sigma$ and 6 for $m$). So the parameter combinations for full factorial design approach are $15 \times 10 \times 6 = 900$ and that for orthogonal experiment design approach might be $15 \times 10 = 150$ (Orthogonal experiment design cannot guarantee to optimal parameter combination as the optimal may be missing in orthogonal design matrix). Second, as ANS is stochastic, it is necessary to run multiple times (e.g. 25 times) to evaluate the effectiveness of each parameter combination. Third, since there are 18 benchmark functions and each function is with 30 and 100 dimensions, the parameter analysis should be performed independently on each function of each specified dimensionality. So even the orthogonal experiment design method is employed, we may need to run ANS at least $15 \times 10 \times 25 = 3750$ times to analyze ANS on one function of one predefined dimensionality.

As the aim is to give a general guidance for the parameter setting of ANS, the parameter sensitivity analysis is not performed rigorously on each function and each dimensionality. Instead, when one parameter is analyzed, other parameters are set to the fixed default values. Users are advised to refer to the parameter analysis results presented in this study and meanwhile employ an advanced technique [33] (e.g. orthogonal experiment design) to configure the parameters of ANS when dealing with a concrete optimization problem.

The results obtained by ANS with different across-search degree values for functions with 30 and 100

dimensions are reported in Table 14 and Table 15, respectively. It can be observed from Table 14 that, as for the benchmark functions with 30 dimensions, the appropriate values of $n$ are usually the end points of the value range (i.e. a set of integers from 1 to the dimensionality value). For example, the best value of $n$ is 1 for functions $f_2$, $f_5$, $f_7$, $f_8$, $f_{10}$, $f_{11}$, $f_{12}$ and $f_{16}$, while the best value of $n$ is 28 (close to the dimensionality value) for functions $f_1$, $f_4$, $f_6$, $f_9$, $f_{13}$, $f_{14}$, $f_{15}$, $f_{17}$ and $f_{18}$. The exception is that when $n$ equals to 10, ANS generates the best results for function $f_3$.

Data in Table 15 show that, the best values of $n$ for functions $f_5$, $f_7$, $f_8$, $f_{10}$, $f_{11}$, $f_{12}$ and $f_{16}$ with 100 dimensions are also 1, which is consistent with the situation of these functions with 30 dimensions. The most appropriate values of $n$ for functions $f_1$, $f_3$, $f_4$, $f_6$, $f_9$, $f_{13}$, $f_{14}$, $f_{15}$, $f_{17}$ and $f_{18}$ with 100 dimensions are also relatively large. One difference is that the best values of $n$ for functions $f_1$, $f_4$, $f_6$, $f_9$, $f_{13}$, $f_{14}$, $f_{15}$, $f_{17}$ and $f_{18}$ with 100 dimensions are among the range from 8 to 20 rather than being close to the dimensionality value (i.e. 100). It is interesting to note that the best value of $n$ for function $f_2$ shifts from 1 to 20 with its dimensionality increasing from 30 to 100.

Based on the observations above, we may conclude that smaller values of $n$ are good for ANS in dealing with the simple multimodal problems. In contrast, when solving the complex rotated problems, larger values of $n$ are usually suggested (except for function $f_{16}$). In addition, appropriate values of $n$ are seemed case-dependent for the simple unimodal problems.

Table 14. Impacts of across-search degree *n* on ANS in solving optimization functions with 30 dimensions. The best results are highlighted.

| Results | *n*=1 | *n*=2 | *n*=4 | *n*=6 | *n*=8 | *n*=10 | *n*=12 | *n*=14 | *n*=16 | *n*=18 | *n*=20 | *n*=22 | *n*=24 | *n*=26 | *n*=28 |
|---|---|---|---|---|---|---|---|---|---|---|---|---|---|---|---|
| $f_1$ | 8.13E-178 | 1.68E-198 | 5.78E-220 | 4.77E-231 | 3.27E-237 | 1.14E-239 | 3.84E-241 | 2.43E-242 | 8.29E-241 | 7.78E-242 | 2.25E-242 | 1.56E-243 | 7.96E-244 | 4.06E-245 | **2.21E-245** |
| $f_2$ | **8.43E+00** | 2.43E+01 | 4.87E+01 | 4.80E+01 | 5.89E+01 | 7.77E+01 | 6.26E+01 | 3.30E+01 | 6.28E+01 | 5.43E+01 | 1.59E+02 | 2.65E+01 | 8.74E+01 | 8.01E+01 | 1.64E+02 |
| $f_3$ | 3.36E-04 | 4.74E-09 | 1.50E-15 | 9.21E-19 | 2.14E-19 | **5.36E-20** | 6.84E-19 | 8.69E-18 | 4.05E-18 | 3.84E-17 | 1.03E-16 | 3.44E-16 | 1.90E-15 | 4.35E-15 | 1.44E-12 |
| $f_4$ | 6.46E-96 | 3.10E-114 | 1.93E-134 | 2.05E-145 | 3.43E-152 | 6.02E-156 | 2.46E-159 | 2.65E-161 | 6.43E-163 | 4.87E-164 | 5.75E-165 | 7.68E-166 | 1.16E-166 | 2.29E-167 | **7.91E-168** |
| $f_5$ | **0.00E+00** | 0.00E+00 | 0.00E+00 | 0.00E+00 | 0.00E+00 | 0.00E+00 | 0.00E+00 | 0.00E+00 | 0.00E+00 | 0.00E+00 | 0.00E+00 | 0.00E+00 | 0.00E+00 | 0.00E+00 | 4.00E-02 |
| $f_6$ | 5.10E-03 | 3.36E-03 | 2.42E-03 | 1.83E-03 | 1.85E-03 | 1.83E-03 | 1.56E-03 | 1.57E-03 | 1.55E-03 | 1.61E-03 | 1.60E-03 | 1.68E-03 | 1.67E-03 | 1.60E-03 | **1.54E-03** |
| $f_7$ | **0.00E+00** | 0.00E+00 | 7.16E-01 | 2.69E+00 | 3.85E+00 | 4.21E+00 | 4.57E+00 | 5.62E+00 | 6.52E+00 | 7.76E+00 | 6.58E+00 | 8.25E+00 | 1.02E+01 | 1.25E+01 | 1.43E+01 |
| $f_8$ | **0.00E+00** | 0.00E+00 | 1.32E+00 | 2.22E+01 | 2.95E+01 | 3.61E+01 | 3.77E+01 | 4.27E+01 | 4.48E+01 | 4.53E+01 | 4.55E+01 | 4.61E+01 | 4.63E+01 | 4.28E+01 | 4.11E+01 |
| $f_9$ | 1.57E-14 | 1.00E-14 | 7.11E-15 | 6.82E-15 | 6.25E-15 | 5.96E-15 | 4.97E-15 | 4.83E-15 | 3.97E-15 | 3.83E-15 | 3.97E-15 | 3.69E-15 | 3.83E-15 | 3.55E-15 | **3.55E-15** |
| $f_{10}$ | **0.00E+00** | 0.00E+00 | 0.00E+00 | 0.00E+00 | 0.00E+00 | 0.00E+00 | 0.00E+00 | 0.00E+00 | 0.00E+00 | 0.00E+00 | 0.00E+00 | 0.00E+00 | 1.08E-03 | 2.95E-04 | 5.91E-04 |
| $f_{11}$ | **1.57E-32** | 1.57E-32 | 1.57E-32 | 1.57E-32 | 1.57E-32 | 1.57E-32 | 1.57E-32 | 1.57E-32 | 1.69E-32 | 2.89E-32 | 3.77E-32 | 8.65E-08 | 2.69E-05 | 4.47E-04 | 2.61E-03 |
| $f_{12}$ | **1.35E-32** | 1.35E-32 | 1.35E-32 | 2.15E-32 | 1.38E-32 | 2.14E-32 | 1.39E-32 | 1.47E-32 | 1.42E-32 | 2.38E-32 | 2.28E-32 | 4.39E-04 | 4.39E-04 | 3.89E-03 | 4.39E-04 |
| $f_{13}$ | 3.03E-90 | 1.05E-126 | 4.27E-161 | 4.24E-175 | 2.11E-182 | 6.47E-187 | 1.34E-190 | 1.68E-192 | 1.01E-193 | 3.50E-195 | 1.25E-196 | 2.91E-198 | 3.14E-198 | 4.57E-199 | **1.71E-199** |
| $f_{14}$ | 2.61E+01 | 2.50E+01 | 2.26E+01 | 2.23E+01 | 2.14E+01 | 2.06E+01 | 2.02E+01 | 2.02E+01 | 2.01E+01 | 1.92E+01 | 1.91E+01 | 1.91E+01 | 2.01E+01 | 1.84E+01 | **1.82E+01** |
| $f_{15}$ | 1.24E-01 | 4.79E-05 | 5.09E-16 | 2.19E-22 | 2.27E-26 | 8.92E-30 | 4.75E-32 | 1.37E-34 | 2.27E-36 | 3.62E-38 | 1.31E-40 | 6.45E-42 | 2.72E-43 | 2.62E-44 | **1.32E-45** |
| $f_{16}$ | **1.61E+02** | 1.47E+02 | 1.70E+02 | 1.73E+02 | 1.74E+02 | 1.77E+02 | 1.75E+02 | 1.71E+02 | 1.71E+02 | 1.73E+02 | 1.72E+02 | 1.67E+02 | 1.67E+02 | 1.73E+02 | 1.68E+02 |
| $f_{17}$ | 1.32E+00 | 1.18E-02 | 7.10E-15 | 6.96E-15 | 6.39E-15 | 5.82E-15 | 4.68E-15 | 4.26E-15 | 3.83E-15 | 3.55E-15 | 3.83E-15 | 3.69E-15 | 3.83E-15 | 3.55E-15 | **3.55E-15** |
| $f_{18}$ | 9.44E-05 | 6.05E-04 | 6.27E-03 | 4.88E-03 | 8.45E-03 | 1.29E-03 | 5.03E-03 | 1.00E-03 | 2.88E-16 | 8.48E-03 | 2.79E-16 | 6.90E-04 | 5.91E-04 | 1.08E-03 | **4.62E-16** |

Table 15. Impacts of across-search degree *n* on ANS in solving optimization functions with 100 dimensions. The best results are highlighted.

| Results | *n*=1 | *n*=2 | *n*=4 | *n*=6 | *n*=8 | *n*=10 | *n*=20 | *n*=30 | *n*=40 | *n*=50 | *n*=60 | *n*=70 | *n*=80 | *n*=90 | *n*=98 |
|---|---|---|---|---|---|---|---|---|---|---|---|---|---|---|---|
| $f_1$ | 2.76E-103 | 9.56E-116 | 5.10E-128 | 6.89E-133 | **8.40E-134** | 1.25E-132 | 4.10E-116 | 3.60E-99 | 6.93E-87 | 6.91E-78 | 1.30E-71 | 2.26E-67 | 3.35E-64 | 4.96E-62 | 4.54E-61 |
| $f_2$ | 9.18E+01 | 8.72E+01 | 9.08E+01 | 8.82E+01 | 8.69E+01 | 8.53E+01 | **8.28E+01** | 8.29E+01 | 8.28E+01 | 8.38E+01 | 8.37E+01 | 8.43E+01 | 8.38E+01 | 8.51E+01 | 8.52E+01 |
| $f_3$ | 5.8665E-01 | 6.5404E-02 | 3.01E-03 | 3.08E-04 | **9.82E-05** | 1.11E-04 | 6.78E-03 | 1.52E-01 | 5.82E-01 | 1.02E+00 | 1.55E+00 | 1.88E+00 | 2.04E+00 | 2.16E+00 | 2.10E+00 |
| $f_4$ | 2.71E-56 | 2.86E-67 | 1.52E-79 | 1.18E-85 | 4.21E-89 | **1.47E-90** | 1.62E-87 | 2.07E-80 | 1.21E-73 | 2.02E-68 | 2.45E-64 | 8.90E-61 | 4.78E-59 | 1.16E-57 | 7.77E-57 |
| $f_5$ | **0.00E+00** | 0.00E+00 | 0.00E+00 | 0.00E+00 | 0.00E+00 | 0.00E+00 | 0.00E+00 | 0.00E+00 | 0.00E+00 | 0.00E+00 | 0.00E+00 | 0.00E+00 | 0.00E+00 | 0.00E+00 | 0.00E+00 |
| $f_6$ | 4.16E-02 | 2.47E-02 | 1.66E-02 | 1.40E-02 | 1.40E-02 | **1.28E-02** | 1.51E-02 | 1.68E-02 | 2.13E-02 | 2.17E-02 | 2.47E-02 | 2.54E-02 | 2.56E-02 | 2.67E-02 | 2.75E-02 |
| $f_7$ | **0.00E+00** | 0.00E+00 | 3.64E+01 | 1.37E+02 | 2.00E+02 | 2.54E+02 | 4.18E+02 | 5.13E+02 | 5.57E+02 | 5.87E+02 | 5.81E+02 | 5.96E+02 | 5.79E+02 | 5.89E+02 | 5.78E+02 |
| $f_8$ | **2.66E-01** | 7.00E-01 | 8.68E+01 | 1.66E+02 | 1.66E+02 | 2.73E+02 | 4.54E+02 | 5.33E+02 | 6.17E+02 | 6.58E+02 | 6.82E+02 | 7.00E+02 | 7.05E+02 | 7.26E+02 | 7.11E+02 |
| $f_9$ | 7.76E-14 | 5.32E-14 | 3.90E-14 | 3.23E-14 | 2.80E-14 | **2.62E-14** | 1.98E-14 | 1.49E-14 | 1.42E-14 | 1.39E-14 | 1.39E-14 | 1.37E-14 | 1.37E-14 | 1.39E-14 | 1.37E-14 |
| $f_{10}$ | **0.00E+00** | 0.00E+00 | 0.00E+00 | 0.00E+00 | 0.00E+00 | 0.00E+00 | 0.00E+00 | 0.00E+00 | 0.00E+00 | 0.00E+00 | 0.00E+00 | 0.00E+00 | 0.00E+00 | 0.00E+00 | 0.00E+00 |
| $f_{11}$ | **1.57E-32** | 1.57E-32 | 1.57E-32 | 1.57E-32 | 1.57E-32 | 1.57E-32 | 1.60E-32 | 1.57E-32 | 1.60E-32 | 1.60E-32 | 1.60E-32 | 1.58E-32 | 6.91E-04 | 1.61E-06 | 2.07E-02 |
| $f_{12}$ | **1.34E-32** | 1.349E-32 | 1.34E-32 | 1.36E-32 | 1.71E-32 | 7.32E-04 | 1.70E-32 | 7.32E-32 | 5.27E-32 | 3.56E-32 | 3.72E-32 | 5.00E-32 | 8.23E-32 | 7.89E-32 | 9.54E-32 |
| $f_{13}$ | 2.88E-103 | 1.04E-115 | 6.29E-128 | 8.51E-133 | 1.23E-133 | **8.01E-133** | 1.76E-116 | 1.62E-99 | 5.57E-87 | 5.63E-78 | 1.78E-71 | 5.89E-67 | 4.60E-64 | 2.08E-62 | 5.59E-61 |
| $f_{14}$ | 9.09E+01 | 9.20E+01 | 9.16E+01 | 9.03E+01 | 8.98E+01 | 8.71E+01 | **8.42E+01** | 8.43E+01 | 8.46E+01 | 8.52E+01 | 8.54E+01 | 8.61E+01 | 8.69E+01 | 8.58E+01 | 8.61E+01 |
| $f_{15}$ | 1.12E-01 | 2.82E-03 | 7.15E-05 | 2.78E-05 | **3.73E-05** | 1.54E-04 | 2.49E-02 | 1.81E-01 | 4.70E-01 | 6.77E-01 | 9.16E-01 | 9.79E-01 | 1.08E+00 | 9.72E-01 | 1.02E+00 |
| $f_{16}$ | **2.17E+02** | 2.94E+02 | 4.49E+02 | 5.39E+02 | 5.95E+02 | 6.45E+02 | 7.30E+02 | 7.72E+02 | 7.87E+02 | 8.23E+02 | 8.12E+02 | 8.21E+02 | 8.17E+02 | 8.23E+02 | 8.22E+02 |
| $f_{17}$ | 3.16E-04 | 5.25E-14 | 3.94E-14 | 3.12E-14 | 2.91E-14 | 2.79E-14 | **1.12E-14** | 1.58E-14 | 1.46E-14 | 1.44E-14 | 1.39E-14 | 1.39E-14 | 1.42E-14 | 1.42E-14 | 1.32E-14 |
| $f_{18}$ | 7.52E-14 | 2.53E-14 | 1.98E-13 | 2.98E-10 | 1.47E-11 | 1.39E-08 | 6.58E-07 | 4.44E-10 | **7.41E-18** | 6.66E-17 | 4.44E-17 | 4.44E-17 | 8.14E-17 | 8.14E-17 | 1.18E-16 |

The results obtained by ANS with different population sizes for benchmark functions with 30 dimensions are listed in Table 16. The data in Table 16 show that it is generally wise to set the population size of ANS to 20 for most of the benchmark functions, except that the best population size for function $f_{16}$ is 30. For all functions, if the population size is too small (less than 5 for example), the performance of ANS degrades rapidly. In addition, for solving simple multimodal functions, it is safe to set the population size to larger values (e.g. equal to or larger than 20). As a result, we may conclude that 20 is a reasonable default value for the population size of ANS when dealing with various types of optimization problems.

Table 16 Impacts of population size *m* on ANS in solving optimization functions with 30 dimensions. The best results are highlighted.

| Results | *m*=5 | *m*=10 | *m*=20 | *m*=30 | *m*=40 | *m*=50 |
|---|---|---|---|---|---|---|
| $f_1$ | 1.34E+05 | 5.69E-07 | **2.21E-245** | 2.95E-153 | 7.67E-111 | 5.62E-86 |
| $f_2$ | 3.28E+01 | 1.26E+01 | **8.43E+00** | 1.16E+01 | 1.22E+01 | 1.26E+01 |
| $f_3$ | 5.20E+00 | 4.37E-01 | **5.36E-20** | 1.48E-15 | 1.55E-11 | 3.29E-09 |
| $f_4$ | 1.70E+01 | 2.67E-05 | **7.91E-168** | 5.74E-107 | 3.68E-78 | 1.19E-61 |
| $f_5$ | 4.57E+01 | **0.00E+00** | **0.00E+00** | **0.00E+00** | **0.00E+00** | **0.00E+00** |
| $f_6$ | 2.10E+01 | 3.01E-02 | **1.54E-03** | 2.01E-03 | 3.11E-03 | 3.47E-03 |
| $f_7$ | 3.38E+00 | **0.00E+00** | **0.00E+00** | **0.00E+00** | **0.00E+00** | **0.00E+00** |
| $f_8$ | 3.09E+03 | 1.80E+00 | **0.00E+00** | **0.00E+00** | **0.00E+00** | **0.00E+00** |
| $f_9$ | 1.21E+01 | 9.31E-02 | **3.55E-15** | **3.55E-15** | **3.55E-15** | **3.55E-15** |
| $f_{10}$ | 2.66E-01 | **0.00E+00** | **0.00E+00** | **0.00E+00** | **0.00E+00** | **0.00E+00** |
| $f_{11}$ | 3.94E-05 | **1.57E-32** | **1.57E-32** | **1.57E-32** | **1.57E-32** | **1.57E-32** |
| $f_{12}$ | 1.36E-02 | **1.35E-32** | **1.35E-32** | **1.35E-32** | **1.35E-32** | **1.35E-32** |
| $f_{13}$ | 2.80E+05 | 1.71E-02 | **1.71E-199** | 1.17E-125 | 2.49E-90 | 6.62E-70 |
| $f_{14}$ | 4.56E+02 | 2.68E+01 | **1.82E+01** | 2.12E+01 | 2.26E+01 | 2.38E+01 |
| $f_{15}$ | 2.38E+00 | 1.41E-01 | **1.32E-45** | 7.89E-30 | 1.37E-22 | 1.84E-17 |
| $f_{16}$ | 2.15E+02 | 1.69E+02 | 1.61E+02 | **1.52E+02** | 1.58E+02 | 1.71E+02 |
| $f_{17}$ | 1.31E+01 | 3.95E-01 | **3.55E-15** | **3.55E-15** | **3.55E-15** | **3.55E-15** |
| $f_{18}$ | 1.37E+02 | 7.37E-02 | **4.62E-16** | 6.66E-16 | 7.91E-15 | 1.39E-15 |

The results of the benchmark functions with 30 dimensions produced by ANS with different values of $\sigma$ are displayed in Table 17. It can be observed that the best values of $\sigma$ are generally among (0.4, 0.6). Especially, ANS with $\sigma = 0.5$ produces either the best or fairly competitive results for all benchmark functions, though ANS with $\sigma = 0.4$ generates the best results for functions $f_1$ and $f_4$ and ANS with $\sigma = 0.6$ obtains the best results for functions $f_2$ and $f_{16}$. In addition, larger values of $\sigma$ are found to be more suitable for ANS when dealing with the simple multimodal functions (e.g. functions $f_7$ - $f_{12}$). Therefore, the promising value range of $\sigma$ should be (0.4, 0.6) and the suggested default value of $\sigma$ should be 0.5, which is consistent with the analysis given in Section 2.2.

Table 17 Impacts of standard deviation $\sigma$ on ANS in solving optimization functions with 30 dimensions. The best results are highlighted.

| Results | $\sigma=0.1$ | $\sigma=0.2$ | $\sigma=0.3$ | $\sigma=0.4$ | $\sigma=0.5$ | $\sigma=0.6$ | $\sigma=0.7$ | $\sigma=0.8$ | $\sigma=0.9$ |
|---|---|---|---|---|---|---|---|---|---|
| $f_1$ | 3.51E+04 | 3.85E+03 | 2.60E+00 | **0.00E+00** | 2.21E-245 | 8.35E-115 | 9.45E-40 | 3.11E-07 | 5.97E+02 |
| $f_2$ | 1.92E+08 | 3.78E+01 | 3.18E+01 | 2.56E+01 | 8.43E+00 | **6.74E+00** | 1.64E+01 | 1.57E+01 | 4.40E+01 |
| $f_3$ | 1.69E+02 | 1.29E+02 | 6.28E+01 | 7.12E-02 | **5.36E-20** | 7.06E-06 | 7.33E-04 | 2.68E+00 | 5.01E+01 |
| $f_4$ | 4.73E+00 | 4.08E-01 | 1.98E-05 | **2.96E-253** | 7.91E-168 | 2.12E-87 | 2.24E-35 | 2.14E-04 | 1.32E-01 |
| $f_5$ | **0.00E+00** | **0.00E+00** | **0.00E+00** | **0.00E+00** | **0.00E+00** | **0.00E+00** | **0.00E+00** | **0.00E+00** | **0.00E+00** |
| $f_6$ | 8.49E+03 | 8.65E+02 | 2.03E+01 | 1.93E-03 | **1.54E-03** | 3.82E-03 | 1.34E-02 | 8.15E-02 | 4.48E+01 |
| $f_7$ | 6.18E+00 | 3.55E-16 | **0.00E+00** | **0.00E+00** | **0.00E+00** | **0.00E+00** | **0.00E+00** | **0.00E+00** | **0.00E+00** |
| $f_8$ | 9.20E+00 | 6.00E-01 | **0.00E+00** | **0.00E+00** | **0.00E+00** | **0.00E+00** | **0.00E+00** | **0.00E+00** | **0.00E+00** |
| $f_9$ | 2.04E+00 | 4.36E-01 | 1.91E-03 | **3.55E-15** | **3.55E-15** | **4.26E-15** | **1.03E-05** | **8.41E-05** | **1.44E-01** |
| $f_{10}$ | 1.40E-01 | **0.00E+00** | **0.00E+00** | **0.00E+00** | **0.00E+00** | **0.00E+00** | **0.00E+00** | **0.00E+00** | **0.00E+00** |
| $f_{11}$ | 1.89E+02 | 4.41E+00 | **1.57E-32** | **1.57E-32** | **1.57E-32** | **1.57E-32** | **1.57E-32** | **1.57E-32** | **1.57E-32** |
| $f_{12}$ | 2.21E+03 | 2.01E-29 | **1.35E-32** | **1.35E-32** | **1.35E-32** | **1.35E-32** | **1.35E-32** | **1.35E-32** | **1.35E-32** |
| $f_{13}$ | 8.50E+04 | 9.29E+03 | 1.38E+02 | 3.90E-210 | **1.71E-199** | 1.68E-93 | 1.99E-29 | 1.99E-29 | 6.18E+03 |
| $f_{14}$ | 3.02E+10 | 6.65E+08 | 2.13E+06 | 7.33E+03 | **1.82E+01** | 2.13E+02 | 4.42E+03 | 2.23E+06 | 5.26E+09 |
| $f_{15}$ | 6.29E+01 | 2.34E+01 | 4.98E+00 | 8.11E-03 | **1.32E-45** | 1.24E-06 | 1.09E-04 | 6.77E+00 | 7.07E+01 |
| $f_{16}$ | 1.97E+02 | 2.01E+02 | 1.69E+02 | 1.69E+02 | 1.61E+02 | **1.58E+02** | 1.67E+02 | 1.66E+02 | 1.69E+02 |
| $f_{17}$ | 2.44E+00 | 1.05E+00 | 1.20E-02 | **3.48E-15** | **3.55E-15** | **3.67E-15** | 6.75E-15 | 7.90E-05 | 1.80E+00 |
| $f_{18}$ | 3.48E-01 | 9.11E-02 | 1.38E-02 | 2.93E-15 | **4.62E-16** | **1.11E-17** | 1.46E-09 | 7.99E-02 | 7.28E-01 |

In conclusion, ANS is sensitive to its parameters. Generally, the required values of across-search degree $n$ may be significantly distinct when ANS is utilized to solve different types of optimization problems. In contrast, the appropriate values of population size $m$ (also the cardinality $c$ of the superior solution set) and standard deviation $\sigma$ are relatively stable for various problems (i.e. $m = 20$ and $\sigma = 0.5$ ). Three steps are recommended to set the parameters of ANS effectively when solving a concrete optimization problem. First, set $m$ and $\sigma$ to their default values 20 and 0.5, respectively. Second, find a proper value for $n$ . Third, do minor modifications to the values of $m$ and $\sigma$ to further improve the performance of ANS.

## 6. Conclusions

A new population-based across neighbourhood search (ANS) is proposed in this study. It is shown that ANS satisfies the following three standards: easy to understand and simple for implementation; unique search strategies different from other PBSAs; competitive performance in dealing with different types optimization problems.

Currently, ANS is in its initial version. To further improve the efficiency of ANS, techniques and ideas from some excellent studies on other population-based search algorithms could be borrowed, such as parameter adaptation, hybridization and coevolution. In addition, the author plans to apply ANS to a number of real-world optimization problems.

**Acknowledgements**

The author thanks Dr. Witold Pedrycz for the supervision in the preparation of this paper. The author thanks Dr. P. N. Suganthan for providing the source codes of the comparative algorithms (jDE, CLPSO, CPSO, FDR and FIPS) and giving insightful suggestion on the experimental design. The author thanks Dr. Jingqiao Zhang for providing the source codes of comparative algorithm JADE. The author thanks Dr. Xin Lv for the insightful discussions on the statistical test. The author thanks the anonymous reviewers for the constructive and helpful comments.